%% file: paper.tex
\documentclass[]{fairmeta}
\usepackage[utf8]{inputenc}
\usepackage[T1]{fontenc}
\usepackage[round]{natbib}
\usepackage{hyperref}
\usepackage{url}
\usepackage{booktabs}
\usepackage{amsfonts}
\usepackage{amsmath}
\usepackage{amssymb}
\usepackage{amsthm}
\usepackage{nicefrac}
\usepackage{microtype}
\usepackage{xcolor}
\usepackage{algorithm}
\usepackage{algorithmic}
\usepackage{graphicx}
\usepackage{lmodern}
\graphicspath{{../experiments/figures/}}
\usepackage{pgfplots}
\pgfplotsset{compat=1.18}
\usepgfplotslibrary{groupplots}
\usepgfplotslibrary{fillbetween}
\usepackage{subcaption}
\usepackage{multirow}
\usepackage{wrapfig}
\input{math_commands.tex}

\newtheorem{theorem}{Theorem}

\newtheorem{proposition}[theorem]{Proposition}

\newtheorem{definition}{Definition}
\newtheorem{assumption}{Assumption}

\newcommand{\care}{\textsc{CARE}}

\newcommand{\cS}{\mathcal{S}}
\newcommand{\cA}{\mathcal{A}}
\newcommand{\cX}{\mathcal{X}}
\newcommand{\cM}{\mathcal{M}}

\newcommand{\cR}{\mathcal{R}}

\newcommand{\cB}{\mathcal{B}}
\newcommand{\cP}{\mathcal{P}}

\newcommand{\cG}{\mathcal{G}}

\title{Self-Improvement Can Self-Regress: The Rise-and-Collapse Failure Mode of LLM Self-Training}

\author[1,*]{Jianzhe Lin}

\affiliation[1]{MetaAI}
\contribution[*]{Work done at Meta}

\abstract{\textbf{Self-improvement can self-regress.} In REINFORCE post-training
for code, a model can improve rapidly on the very metric it optimizes
and then collapse within the same training campaign. We document this
failure mode in a controlled, multi-seed testbed: Qwen-2.5 at 3B and 7B,
trained on competitive-programming tasks with binary CodeGrader reward
over 10 sequential 20-step campaigns. Across campaigns, pass@1
exhibits a robust \emph{rise-then-collapse}: it climbs to a peak in
tens of gradient steps and then falls back, sometimes to near zero.
This is not catastrophic forgetting across tasks; it is within-task
policy over-optimization on a fixed distribution, and KL- and
EWC-style parameter-level constraints do not prevent it.

The operational question is then: \emph{where should the control loop
sit?} We compare three intervention levels on the same testbed with
bootstrap 95\% CIs. \textbf{\care{}} is a deliberately specified
stress test of \emph{between}-campaign memory: a capability posterior,
multi-action transfer gate, and regression-aware belief revision.
\textbf{ES} is a deployed \emph{within}-campaign early-stop rule that
rolls forward the peak checkpoint and sets the next campaign's budget
to $\text{peak\_step}{+}3$. \textbf{GRPO} changes the underlying RL update rule, using
group-relative reward normalization rather than vanilla REINFORCE.

The answer is \textbf{regime-dependent}. At \textbf{Qwen-2.5-3B},
where naive REINFORCE is fragile, \care{} v2 nearly doubles mean
end-of-chain pass@1 from $4.9\%$ to $\mathbf{9.5\%}$ (paired bootstrap
95\% CI of the per-seed difference $[+0.4, +8.9]$, excludes zero,
$n{=}5$, positive in $4/5$ seeds); it also beats ES ($8.2\%$, $n{=}5$).
At \textbf{Qwen-2.5-7B}, \care{} reaches parity with naive REINFORCE
($13.8\%$ vs $11.8\%$), while ES reaches $\mathbf{22.2\%}$
[14.1, 28.0] ($n{=}3$). However, out-of-the-box \textbf{GRPO} reaches
$\mathbf{20.7\%}$ [15.7, 25.1] ($n{=}5$) without any campaign-level
orchestration, nearly matching orchestrated REINFORCE+ES; adding
\care{} on top of GRPO remains at parity ($20.4\%$ [10.9, 28.6],
$n{=}5$).

\textbf{GRPO raises the floor, but does not remove the cliff.}
Per-step trajectory diagnostics show that GRPO's 7B gain comes from
improved \emph{between-campaign} carryover, not from within-campaign
stabilisation: the per-campaign peak-to-end gap is $\approx 17$ pt
under both REINFORCE and GRPO. A deployed GRPO+ES run gives mixed
evidence: 2/3 seeds improve over naive GRPO, but one final-campaign
cliff drives the 3-seed mean to $17.0\%$ [0.0, 28.1], showing that
even algorithm-level and within-campaign control together do not
eliminate single-campaign collapse. A single-seed Gemma-3-4B pilot
shows the same qualitative rise-then-collapse signature on the
identical CodeGrader testbed (peak $32.8\%$, end $0\%$); we report
this as a pilot data point, not a multi-seed replication, but it
suggests the phenomenon is not confined to the Qwen family.}

\date{\today}
\correspondence{Jianzhe Lin at \email{jianzhelin@meta.com}}

\begin{document}

\maketitle

%==============================================================================
\section{Introduction}
\label{sec:intro}
%==============================================================================

Self-improving AI systems are proliferating: iterative DPO~\citep{xu2024iterativedpo, pang2024iterative}, self-play~\citep{singh2024selfplay, chen2024selfplay}, self-reward training~\citep{yuan2024selfreward}, STaR~\citep{zelikman2022star}, and autonomous research agents~\citep{lu2024aiscientist} all aim to make models better by learning from their own outputs. The promise is a virtuous cycle: the model improves, generates better data, and improves further.

But self-improvement is not a free lunch. Concurrent work on RLVR (reinforcement learning with verifiable rewards) has begun to surface a closely related failure mode: policy entropy collapses, exploration shrinks, and the same RL update that initially improved the model later destroys it~\citep{prosperity2026collapse, flexible2026entropy, exploration2026rlvr}. We document the same effect from a different angle --- a within-campaign \emph{rise-then-collapse} on a binary CodeGrader reward --- and ask which level of intervention actually moves end-of-chain pass@1 once it appears.

Consider a concrete failure mode. An autonomous research agent discovers that increasing the rejection sampling ratio from 4$\times$ to 16$\times$ improves pass@1 on code generation by 6\%. Encouraged, it applies the same strategy to the next campaign. But this time, the strategy \emph{also} reduces solution diversity by 4\% and degrades hard-case accuracy by 1\%---regressions invisible to the agent because it only tracked the target metric. Over repeated campaigns, the model becomes narrowly optimized: high on its target metric but brittle, homogeneous, and fragile on distribution shifts.

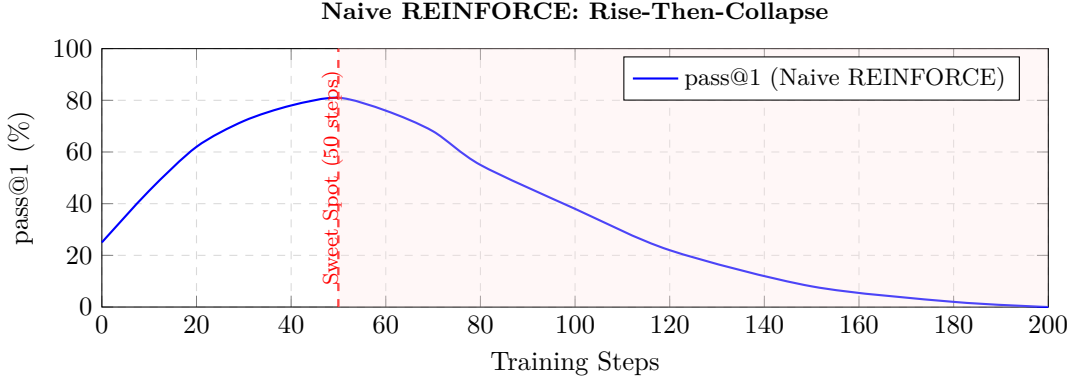
\begin{figure}[!t]
\centering
\begin{tikzpicture}
\begin{axis}[
    width=0.85\columnwidth,
    height=5cm,
    xlabel={Training Steps},
    ylabel={pass@1 (\%)},
    xmin=0, xmax=200,
    ymin=0, ymax=100,
    grid=major,
    grid style={dashed, gray!30},
    legend pos=north east,
    legend style={font=\small},
    title style={font=\small\bfseries},
    title={Naive REINFORCE: Rise-Then-Collapse},
]
% Empirical data from care_measy_lr1e6_kl_full_200steps-b77ssz9h
\addplot[blue, thick, mark=none, smooth] coordinates {
    (0, 25) (10, 45) (20, 62) (30, 72) (40, 78) (50, 81)
    (60, 76) (70, 68) (80, 55) (100, 38) (120, 22) (150, 8) (180, 2) (200, 0)
};
\addlegendentry{pass@1 (Naive REINFORCE)}

% Sweet spot annotation
\draw[red, dashed, thick] (axis cs:50,0) -- (axis cs:50,100);
\node[red, font=\footnotesize, rotate=90, anchor=south] at (axis cs:53,50) {Sweet Spot (50 steps)};

% Collapse region
\fill[red!10, opacity=0.3] (axis cs:50,0) rectangle (axis cs:200,100);
\node[red!60!black, font=\footnotesize] at (axis cs:130,85) {Collapse Region};

\end{axis}
\end{tikzpicture}
\caption{\textbf{Rise-then-collapse in naive REINFORCE training} (Qwen-2.5-7B on competitive programming): pass@1 rises from 25\% to 81\% in the first 50 steps, then degrades to near-zero by step 200. Self-improvement is also, on its own target metric, self-regression. \emph{This is a diagnostic single 200-step run from one seed used to characterise the shape of the trajectory; the same rise-then-collapse pattern is shown across multiple seeds and both 3B and 7B scales in Figure~\ref{fig:rise_collapse_chain} (chained 20-step campaigns) and quantified in Section~\ref{sec:discussion} (phase-transition score, onset step, post-onset latency).} The remainder of the paper compares interventions \emph{between} campaigns (\care{}), \emph{within} a campaign (deployed early-stop, ES), and at the \emph{algorithm level} (GRPO; Section~\ref{sec:exp_grpo}) for keeping end-of-chain pass@1 above this collapse.}
\label{fig:rise_collapse}
\end{figure}

This is not a hypothetical. We observe it directly in our REINFORCE training experiments (Figure~\ref{fig:rise_collapse}): when training Qwen-2.5-7B on competitive programming tasks, pass@1 rises from 25\% to 81\% within the first 50 gradient steps---then \emph{collapses to near-zero by step 200}. The model improves rapidly, then catastrophically forgets how to generate correct code. This ``rise-then-collapse'' pattern is the within-campaign manifestation of the same phenomenon: naive optimization without memory or constraint accumulates hidden damage that eventually overwhelms the gains.

\paragraph{What the collapse is, and is not.} The collapse is \emph{not} primarily caused by task switching. It occurs within a nominally fixed competitive-programming training distribution, with no change of dataset or objective. We interpret it as \emph{within-task policy over-optimization}: early REINFORCE sharpens useful behaviours already present in the pretrained model, but continued optimization narrows the policy around brittle reward-correlated patterns, reducing solution diversity and overwriting broad code-generation priors. The model can self-regress without any task change. This distinguishes our setting from the catastrophic-forgetting literature (where forgetting is induced by switching tasks), and the magnitude of each campaign's collapse---together with how much of the within-campaign peak the chain retains---turns out to be scale-dependent. Sections~\ref{sec:experiments}--\ref{sec:discussion} measure that scale dependence and trace it to three structural properties of the per-step trajectory.

More broadly, we show empirically (Section~\ref{sec:multicap}) that naive self-improvement pipelines consistently improve their target metric while causing hidden regressions on secondary capabilities---a phenomenon we call \textbf{capability regression through scalar optimization}.

\paragraph{The Core Insight.} We observe that improvement strategies are not scalar operators (``+0.06 on pass@1'') but \emph{capability tradeoff operators}---they shift an entire vector of capabilities:
\begin{equation}
    s: \mathbf{c} \mapsto \mathbf{c} + \boldsymbol{\delta}_s(\mathbf{c}, \text{ctx})
\end{equation}
where $\mathbf{c}$ is a $K$-dimensional capability vector and $\boldsymbol{\delta}_s$ is the strategy's \emph{capability effect}, which depends on the current capability state and context. A self-improving system that remembers only ``this strategy improved the score'' is discarding most of the information it needs to improve safely and cumulatively.

\paragraph{Our Approach: \care{}.} We introduce \textbf{C}apability-\textbf{A}ware \textbf{R}esearch \textbf{E}xperience (\care{}), a meta-scientific memory system with three modules:
\begin{enumerate}
    \item \textbf{Capability-Effect Memory} records each strategy's full capability delta, boundary conditions, and confidence.
    \item \textbf{Self-Improvement Transfer Gate} decides whether to reuse, adapt, pilot, or reject a past strategy---maximizing target improvement subject to no hidden regression on protected capabilities.
    \item \textbf{Regression-Aware Belief Revision} updates memory when a strategy's observed effects violate predictions, enabling the system's self-improvement policy to itself improve over time.
\end{enumerate}

The paper's central question is operational: \textbf{where should the control loop sit} when a self-improving REINFORCE run keeps collapsing? We study three levels of intervention at three different timescales. \textbf{Between-campaign} control (\care{}): after each campaign, decide what to keep and what to roll back via a capability-aware memory and gate. \textbf{Within-campaign} control (ES): during each campaign, watch the per-step trajectory and stop in time. \textbf{Algorithm-level} control (GRPO~\citep{shao2024deepseekmath, guo2025deepseekr1}): change the RL update itself, so the underlying loss is less prone to collapse in the first place. These act at different timescales (after a campaign, during a campaign, and inside each gradient step) and exploit different signals; we ask which one actually moves end-of-chain pass@1 in our testbed.

\paragraph{Contributions.}
\begin{enumerate}
    \item \textbf{Rise-then-collapse under verifiable-reward REINFORCE.} On Qwen-2.5 3B and 7B trained with REINFORCE and binary CodeGrader reward, pass@1 rises to a peak in tens of gradient steps and then collapses within the same campaign (Figure~\ref{fig:rise_collapse}); chained over 10 campaigns the pattern repeats across multiple seeds and both scales (Figure~\ref{fig:rise_collapse_chain}). Parameter-level regularization (EWC-style KL constraints) does not prevent it (Table~\ref{tab:prelim_kl}).
    \item \textbf{Between-campaign memory has a fragile-regime niche.} At Qwen-2.5-3B, \care{} v2 nearly doubles mean end-of-chain pass@1 ($\mathbf{9.5\%}$ vs $4.9\%$; paired bootstrap 95\% CI of per-seed difference $[+0.4, +8.9]$, excludes zero, $n{=}5$; Table~\ref{tab:care_longchain}) \emph{and} beats a deployed within-campaign early-stop rule (ES, $8.2\%$, $n{=}5$). At Qwen-2.5-7B \care{} reaches parity with naive ($13.8\%$ vs $11.8\%$, overlapping CIs and paired CI includes zero); the value of between-campaign memory is bounded to the small-model fragile-signal regime.
    \item \textbf{Within-campaign stopping sets a higher 7B ceiling under REINFORCE.} The ES condition ($\text{max\_steps}$ $(c{+}1) = \text{peak\_step}(c) + 3$, base hyperparameters preserved, peak checkpoint rolled forward) reaches $\mathbf{22.2\%}$ [14.1, 28.0] end pass@1 on Qwen-2.5-7B 10-campaign REINFORCE chains ($n{=}3$, Table~\ref{tab:care_oracles}), the highest end-of-chain pass@1 of any orchestrated REINFORCE configuration we tested at this scale.
    \item \textbf{GRPO raises the end-of-chain floor without removing the within-campaign cliff.} Replacing REINFORCE with GRPO (group-relative reward normalization, no algorithmic change to the orchestrator) under \emph{the same} 10-campaign 20-step chain protocol yields $\mathbf{20.7\%}$ [15.7, 25.1] ($n{=}5$) at 7B from naive GRPO alone, nearly matching REINFORCE+ES (Table~\ref{tab:grpo_central}). However, per-step trajectory diagnostics on the W4 vs W17 7B A0 chains (Table~\ref{tab:grpo_traj}) show that GRPO does \emph{not} close the within-campaign peak-to-end gap: both update rules leave $\approx 17$ pt of peak-end gap per campaign. GRPO's gain instead comes from better \emph{between-campaign} carryover (mean end pass@1 stays at the 20\% level across the chain, while REINFORCE's drifts down). \care{} on top of GRPO gives no measurable improvement at $n{=}5$. This suggests a natural complementarity hypothesis between GRPO and ES (ES still has the within-campaign peak left to recover under GRPO), which W18 (GRPO+ES, $n{=}3$) only partially supports: 2/3 seeds improve over naive GRPO, but a single final-campaign cliff prevents a statistically separated mean gain.
    \item \textbf{A two-timescale account of where each level helps.} Self-improving RL on this testbed fails at two distinct timescales --- \emph{within} a campaign (cliff-like rise-then-collapse: phase-transition score $\approx 0.78$, onset near step $\approx 17/20$, zero usable post-onset latency for an end-of-campaign gate; Section~\ref{sec:discussion}) and \emph{between} campaigns (each campaign's end checkpoint either carries forward gain or drifts down). The three intervention levels target different cells of this $(within, between)$ decomposition: \care{} helps fragile carryover at 3B; ES recovers within-campaign peaks under REINFORCE; GRPO improves carryover at 7B but leaves the within-campaign cliff approximately intact.
\end{enumerate}

\paragraph{Positioning relative to prior work.} Prior work identifies
\emph{what} self-improvement degrades --- diversity, OOD
generalization, prompt-following, and exploration
\citep{wu2024selfimprovementreversal, kumar2025selftrain,
chen2025passk, kumar2024score, singhal2025rewarddesign}. We instead
study \emph{when} degradation becomes actionable inside a chained RL
training loop. This temporal structure determines the correct
control level: between-campaign memory when the per-step signal is
weak, and within-campaign stopping when it is rich. We are explicitly
\emph{not} claiming to be the first to observe collapse or
self-regression in self-improving LLM training.

%==============================================================================
\section{Related Work}
\label{sec:related}
%==============================================================================

\paragraph{Self-Improving Language Models.}
Self-play~\citep{singh2024selfplay, chen2024selfplay}, self-reward~\citep{yuan2024selfreward}, iterative DPO~\citep{xu2024iterativedpo, pang2024iterative}, STaR~\citep{zelikman2022star}, and ReST~\citep{gulcehre2023rest} optimize a single target metric without tracking capability tradeoffs. SPIN~\citep{chen2024selfplay} frames self-improvement as a game but still uses scalar reward. These methods aim to make self-improvement compound; the question we take up is when and where this compounding fails inside a chained RL training loop, and which intervention level can stop the failure (Section~\ref{sec:experiments}).

\paragraph{Self-Improvement Reversal, Reward Over-Optimization, and Exploration Collapse.}
A growing line of work documents that self-improvement under scalar or
self-generated feedback degrades non-target capabilities even when the
target metric continues to rise.
\citet{wu2024selfimprovementreversal} call this
\emph{self-improvement reversal}: iterative post-training (SFT, DPO,
SFT$\rightarrow$DPO) raises pass@1 while reducing output diversity and
out-of-distribution generalization, and biases the model toward easier
problems rather than genuinely harder ones.
\citet{kumar2025selftrain} report \emph{template collapse} under
prolonged self-consistency-based self-training: large reasoning models
learn to ignore the prompt and emit a fixed template answer, which
they attribute to the self-reinforcing nature of imperfect feedback.
\citet{chen2025passk} observe that Pass@1 reward in RLVR pushes the
policy toward conservative, similar actions and suppresses
exploration, motivating Pass@$k$-style reward redesigns that
explicitly balance exploration vs.\ exploitation.
\citet{kumar2024score} report behaviour collapse during self-correction
RL and prescribe a regularised two-stage training procedure (separating
the initialisation that learns a correction strategy from the
high-reward fitting that would otherwise overwrite it).
\citet{singhal2025rewarddesign} study binary vs.\ pass-rate reward
design for code-generation RL and find that pass-rate's early
advantage typically disappears with continued training, suggesting
that the failure mode is at the level of gradient \emph{direction}
rather than reward \emph{density}.
Concurrent 2026 work on RLVR makes the same diagnosis at the
algorithm level: \citet{prosperity2026collapse} characterise
``prosperity-before-collapse'' under off-policy RLVR with stale data
and trace it to entropy collapse;
\citet{flexible2026entropy} propose a gradient-preserving entropy
controller that targets the same collapse phenomenon; and
\citet{exploration2026rlvr} reframe RLVR's clipping schedule as an
exploration--exploitation knob. These works treat the cliff as an
algorithm-level pathology to be fixed by changing the update rule.
We complement that line by treating the cliff as a fixed empirical
phenomenon and asking which level of \emph{control} around the rule
actually recovers end-of-chain pass@1.
\citet{unsupervised2026rlvr} push RLVR scaling without external
labels; whether the rise-then-collapse signature we observe also
appears in their unsupervised regime is a natural follow-up.
Prior work identifies \emph{what} self-improvement degrades:
diversity, OOD generalization, prompt-following, and exploration. We
instead study \emph{when} the degradation becomes actionable inside a
chained RL training loop, and show that this temporal structure
determines the correct control level --- between-campaign memory when
the per-step signal is weak, within-campaign stopping when it is rich
(Sections~\ref{sec:experiments}--\ref{sec:discussion}).

\paragraph{Autonomous Research Agents.}
The AI Scientist~\citep{lu2024aiscientist} and FunSearch~\citep{romera2024funsearch} automate experimentation, but treat each experiment independently. They rediscover which strategies work without building persistent memory of \emph{how strategies trade off capabilities}. \care{} tests whether the meta-scientific memory absent in these agents would help; our results show this memory is genuinely useful in the small-model regime (Qwen-2.5-3B) but redundant once the per-campaign signal is strong (Qwen-2.5-7B), and an online within-campaign rule dominates at both scales.

\paragraph{Hyperparameter and Strategy Optimization.}
Bayesian optimization~\citep{snoek2012practical, balandat2020botorch} and population-based training~\citep{jaderberg2017pbt} search for good configurations but optimize a scalar objective. AgentHPO~\citep{liu2024agenthpo} uses LLMs to reason about hyperparameters but lacks structured memory of capability effects. Multi-objective HPO~\citep{daulton2020morbo, karl2022mobo} considers multiple objectives simultaneously but does not learn transferable tradeoff knowledge.

\paragraph{Continual Learning and Catastrophic Forgetting.}
EWC~\citep{kirkpatrick2017ewc}, SI~\citep{zenke2017synaptic}, and GEM~\citep{lopez2017gem} address forgetting of \emph{task knowledge}---the model forgets how to do task A after learning task B. Our setting is distinct: we observe within-task self-regression under continued REINFORCE on a fixed distribution (Section~\ref{sec:intro}, ``What the collapse is''), and we test whether a meta-level memory of safe-vs-unsafe improvement strategies can prevent it. Our scale-dependent result (Sections~\ref{sec:experiments}--\ref{sec:discussion}) is consistent with online within-campaign prediction being the more general ingredient, with campaign-level memory acting as a useful but scale-bounded complement.

\paragraph{Meta-Learning.}
MAML~\citep{finn2017maml} learns initializations for fast adaptation; Meta-SGD~\citep{li2017metasgd} learns per-parameter rates. These meta-learn over model parameters. \care{} is positioned as meta-learning over the \emph{self-improvement process}---which interventions improve which capabilities and which they regress; whether this is the right level of abstraction is the question we test empirically.

\paragraph{Curriculum Learning and Data Selection.}
Curriculum learning~\citep{bengio2009curriculum}, self-paced learning~\citep{kumar2010selfpaced}, and data mixing~\citep{xie2024doremi} select what to train on but treat curriculum effects as scalar improvements. \care{} models curriculum strategies as capability tradeoff operators that may help some capabilities while hurting others.

%==============================================================================
\section{Problem: Self-Improvement is Not a Free Lunch}
\label{sec:problem}
%==============================================================================

A model's \emph{capability state} is a vector
$\mathbf{c} = (c_1, \ldots, c_K) \in [0,1]^K$ measuring performance on $K$
held-out evaluation axes (e.g., matched-difficulty pass@1, held-out hard
pass@1, OOD pass@1). A training strategy $s$ acts as a capability
operator: applying $s$ in context $x \in \cX$ yields a capability delta
$\boldsymbol{\delta}_s(x) \in R^K$ so that $\mathbf{c}' = \mathbf{c} +
\boldsymbol{\delta}_s(x)$. We say $s$ causes \emph{hidden regression} on
a protected capability $j \in \cP \subseteq [K]$ if
$\delta_s^{k^*} > 0$ but $\delta_s^j < -\epsilon$ for the target
capability $k^*$ and a threshold $\epsilon$; that is, the strategy
improves the target while silently degrading a protected dimension.
Scalar-objective self-improvement accepts any $s$ with
$\delta_s^{k^*} > 0$ regardless of $\boldsymbol{\delta}_s$ on protected
axes, so under any distribution of strategies for which
$E[\delta_s^j \mid \delta_s^{k^*} > 0] < 0$ on protected $j$, the
expected protected-capability drift after $T$ campaigns is $-\mu_j T$.
We confirm this pattern empirically in Section~\ref{sec:multicap}: scalar
pass@1 improvements on the training distribution do not transfer to
held-out hard or non-English splits regardless of training method.
\emph{Negative transfer} is the additional risk that a strategy
beneficial in one context $x_1$ becomes harmful in another $x_2$, which
a memory-free chain cannot detect. Formal definitions
(Capability Vector, Training Strategy as Capability Operator, Hidden
Regression, Regression Accumulation, Negative Transfer) are in
Appendix~\ref{app:problem_full}.

%==============================================================================
\section{\care{} as a Diagnostic Foil for Campaign-Level Control}
\label{sec:method}
%==============================================================================

\paragraph{What \care{} is, and is not, in this paper.} \care{} is
\emph{not} claimed here as a complete $K$-dimensional deployed
controller for self-improving LLM training. It is a deliberately
specified \emph{stress test of the campaign-level intervention class}:
the K-dimensional capability vector, Gaussian posterior, and gate
routing are written down in full so that ``campaign-level memory + gate
+ belief revision'' is a concrete, falsifiable object that subsequent
empirical sections can interrogate. The deployed gate in our headline
experiments uses only the scalar end/peak pass@1 ratio (i.e.\ $K{=}1$;
see Section~\ref{sec:multicap}); the multi-capability formalism below
motivates the broader design space and is audited post-hoc, not
asserted as a fielded result. Readers who prefer to skip the
mathematical specification can read Modules~1--3 below at the level of
the architecture box (Figure~\ref{fig:architecture}) and then proceed
directly to the experimental sections.

\care{} consists of three tightly coupled modules that together enable safe, cumulative self-improvement (Figure~\ref{fig:architecture}).

\begin{figure}[t]
\centering
\fbox{\parbox{0.95\textwidth}{
\textbf{Module 1: Capability-Effect Memory} $\cM$ \\
\hspace*{1em}Records structured entries: $\{s, \text{ctx}, \boldsymbol{\delta}_s, \text{boundary}, \text{confidence}\}$ \\[4pt]
\textbf{Module 2: Self-Improvement Transfer Gate} $\cG$ \\
\hspace*{1em}Input: candidate strategy $s$, current context $x$, protected caps $\cP$ \\
\hspace*{1em}Output: $\texttt{reuse} \mid \texttt{adapt} \mid \texttt{pilot} \mid \texttt{reject}$ \\
\hspace*{1em}Objective: $\max \E[\delta_s^{k^*}]$ s.t. $\Pr[\delta_s^j < -\epsilon] \leq \alpha, \; \forall j \in \cP$ \\[4pt]
\textbf{Module 3: Regression-Aware Belief Revision} $\cR$ \\
\hspace*{1em}Trigger: observed $\boldsymbol{\delta}_s^{\text{obs}}$ deviates from predicted $\hat{\boldsymbol{\delta}}_s$ \\
\hspace*{1em}Action: update boundary conditions, confidence, and transfer policy
}
}
\caption{\care{} architecture. Module 1 stores capability-level effects. Module 2 gates strategy transfer. Module 3 revises beliefs when predictions fail. Together they make self-improvement cumulative and regression-aware.}
\label{fig:architecture}
\end{figure}

\subsection{Design Sketch}
\label{sec:design_sketch}

\care{} is the natural memory-and-gating candidate for campaign-level
intervention against self-regression. Below we sketch the three modules at
the level of detail needed to interpret the experiments; the full
mathematical specification (entry tuple, posterior update, gate routing
logic, belief-revision Bayesian update, and Algorithm~1 pseudocode) is in
Appendix~\ref{app:care_full}.

\paragraph{Module 1: Capability-Effect Memory $\cM$.} Stores one entry per
(strategy, context) pair containing the observed capability delta
$\boldsymbol{\delta}_s^{\text{obs}} \in R^K$, a boundary predicate over
contexts where the strategy is judged safe, and a confidence score. Multiple
observations of the same strategy are aggregated as a linear Gaussian
posterior over $\boldsymbol{\delta}_s(x)$ conditioned on context features
$\phi(x)$. The intent is a transferable record of \emph{when} a strategy
helps or hurts, not just \emph{whether} it improved a scalar metric.

\paragraph{Module 2: Self-Improvement Transfer Gate $\cG$.} Given a candidate
strategy $s$ and current context $x$, the gate selects one of four actions
$\{\texttt{reuse}, \texttt{adapt}, \texttt{pilot}, \texttt{reject}\}$ to
maximise expected target improvement subject to a per-protected-capability
regression-probability constraint. Concretely, we evaluate
$\Pr[\delta_s^j < -\epsilon \mid x]$ via the posterior CDF for each
protected $j$ and reject any strategy whose worst-case
regression probability exceeds a threshold $\alpha$. In the experiments we
implement a simplified hard-rule version (deterministic ratio test on
observed end/peak pass-rate). Critically for our negative result: this
signal is only computable \emph{at end of campaign}, not during it.

\paragraph{Module 3: Regression-Aware Belief Revision $\cR$.} When the
observed capability delta deviates from the posterior prediction by more
than a Mahalanobis-distance threshold, the system simultaneously refines
the strategy's boundary predicate, updates the Gaussian posterior with
inflated noise, and tightens the gate's regression-probability threshold
$\alpha$. The intent is that the orchestrator's transfer policy
\emph{itself} improves over time as it accumulates surprise events.

\paragraph{Conceptual analysis as a \emph{diagnostic foil}, not a
guarantee (full version in Appendix~\ref{app:care_full}).} Under
idealising assumptions (Lipschitz capability effects, $\sigma_0/\sqrt{m}$
posterior coverage) the long-run hidden regression rate under \care{}
would be $\alpha + O(K/\sqrt{T})$ across $T$ campaigns vs.\ $\Theta(1)$
for a scalar chain, and belief-revision accuracy would grow monotonically
with the number of memory entries. We include the analysis not as a
deployment guarantee but as a \emph{foil}: it formalises the smooth
regime in which campaign-level memory would be expected to help, and the
experiments in Section~\ref{sec:discussion} show exactly why REINFORCE
LLM training violates that regime (phase-transition score $\approx 0.78$,
zero usable post-onset latency). Whether the abstraction holds is itself
part of the empirical contribution.

\paragraph{Design intent vs.\ empirical outcome.} \care{} is not
proposed here as a universally best intervention. It is a deliberately
plausible campaign-level candidate used to test whether memory and
gating are the right level of control for self-improvement collapse.
Section~\ref{sec:experiments} reports a scale-dependent answer:
\care{} \emph{significantly outperforms naive in the Qwen-2.5-3B fragile
regime} ($9.5\%$ vs $4.9\%$; paired bootstrap 95\% CI of per-seed difference $[+0.4, +8.9]$ excludes zero, $n{=}5$) but
\emph{does not significantly outperform naive at Qwen-2.5-7B} ($13.8\%$
vs $11.8\%$, overlapping CIs); at both scales the deployed within-campaign
ES rule (Section~\ref{sec:exp_oracles}) reaches a higher ceiling.
Section~\ref{sec:discussion} diagnoses why.

\paragraph{Scope of the gate's signal in this paper.} Our deployed
\care{} implementation uses a \emph{scalar} gate over the end/peak pass@1
ratio (i.e.\ $K{=}1$ at training time); the multi-capability vector
formalism above motivates the broader class but the cross-language
(\texttt{cpp\_acc}) and OOD (\texttt{ood\_acc}) measurements in
Section~\ref{sec:multicap} enter only as post-hoc diagnostic
evaluations, not as control signals. A post-hoc K-dim audit
(Table~\ref{tab:care_kdim}) shows the multi-capability information is
non-trivial; extending the deployed gate to a true K-dim posterior is
left as future work.

\paragraph{Honest framing of ``capability-aware''.} The word
\emph{capability-aware} in \care{}'s name and Section~\ref{sec:intro}
contribution list refers to the framework's \emph{design space} ---
the K-dimensional capability operator, the protected-set objective,
the Gaussian posterior in Appendix~\ref{app:care_full} --- not to the
deployed gate used in this paper's headline experiments. What is
deployed is the scalar reduction $K{=}1$ on pass@1, plus the
multi-action transfer gate (\texttt{reuse}/\texttt{adapt}/\texttt{pilot}/\texttt{reject})
and the regression-aware belief revision, both of which act on that
scalar signal. Readers should therefore interpret our 3B positive
trend (Section~\ref{sec:exp_longchain}) as evidence that a
\emph{scalar-gated} version of \care{}, not a fully K-dimensional one,
provides the small-model benefit. Whether activating the K-dim
posterior would tighten the 3B effect or extend it to 7B is the
single most informative follow-up experiment to the present paper,
and we discuss it in the exploratory K-dim audit
(Appendix~\ref{app:exploratory}) rather than over-claim it in the
main results.

%==============================================================================
\section{Experiments}
\label{sec:experiments}
%==============================================================================

\subsection{Setup}
\label{sec:setup}

\paragraph{Testbed.} Code generation self-improvement across 5 sequential campaigns: (C1) String Manipulation, (C2) Basic Data Structures, (C3) Advanced Data Structures, (C4) Algorithm Design, (C5) Systems-Level. Problems drawn from HumanEval~\citep{chen2021humaneval}, MBPP~\citep{austin2021mbpp}, and APPS~\citep{hendrycks2021apps}. Each campaign: 50 training problems, 30 held-out evaluation problems.

\paragraph{Campaign Length: three distinct settings.} The paper uses
three deliberately-chosen campaign-length protocols. \textbf{(1)
Diagnostic 200-step run} (Figure~\ref{fig:rise_collapse}, single seed,
Qwen-2.5-7B): we train continuously for 200 gradient steps without
chaining to characterise the rise-then-collapse trajectory --- pass@1
rises from $25\%$ to $81\%$ within the first $\approx 50$ steps and then
collapses to near-zero by step $200$. This is the only place the
$200$-step number appears. \textbf{(2) Module-ablation
campaigns (50 steps $\times$ 3 sequential campaigns $\times$ 5 seeds;
Wave~1v2, Section~\ref{sec:exp_ablation},
Table~\ref{tab:care_ablation_v2})}: longer per-campaign budget chosen
to give the gate's transfer logic more signal per decision while still
chaining only a few times. \textbf{(3) Headline long-chain protocol
(20 steps $\times$ 10 sequential campaigns; Waves~4, 7, 8, 12, 14, 15,
16; Sections~\ref{sec:exp_longchain}--\ref{sec:exp_oracles})}: short
per-campaign budget chained over 10 campaigns to maximise the
between-campaign decision count where \care{}'s memory and gate are
exercised. All headline scale comparisons, ES deployments, and
oracle/baseline numbers reported as our main results use this $20\times
10$ protocol. Per-protocol step counts and chain lengths are repeated
explicitly in each section's table caption so the reader never has to
infer which setting a number came from.

\paragraph{Capability Vector.} In the current experiments we track a single primary capability:
\begin{itemize}
    \item \textbf{pass@1}: functional correctness on competitive programming (CodeGrader execution)
\end{itemize}
Multi-dimensional capability tracking (diversity, hard-case accuracy, robustness, generalization) is the target of ongoing experiments.

\paragraph{Base Models.} \textbf{Qwen-2.5-3B-Instruct} and \textbf{Qwen-2.5-7B-Instruct} for the headline scale comparisons in Section~\ref{sec:exp_longchain} (Waves~4/7/8/12/14/15/16); \textbf{Qwen-2.5-32B-Instruct} is used only as a frozen reference baseline (no training, Section~\ref{sec:exp_scale}, Table~\ref{tab:care_scale}). All training uses REINFORCE on competitive programming problems with CodeGrader as the binary reward.

\paragraph{Why Qwen-2.5 (and not the newest Qwen generation).} We use Qwen-2.5 rather than the newest Qwen generation because it provides a stable, widely used, open-weight model family with dense 3B/7B checkpoints, allowing controlled scale comparisons under the same training stack. Our goal is not to benchmark the latest Qwen model, but to study the trajectory geometry of self-improving RL. We therefore complement the Qwen-2.5 study with a single-seed Gemma-3-4B pilot (Section~\ref{sec:limitations}) as an initial cross-family check.

\paragraph{Baselines.}
\begin{itemize}
    \item \textbf{Naive (no KL)}: Standard REINFORCE with no regularization constraint.
    \item \textbf{EWC (KL=0.05)}: Strong KL penalty to reference policy (EWC-style constraint).
    \item \textbf{Adaptive KL (0.01$\to$0.02$\to$0.03)}: Progressively increasing KL schedule.
\end{itemize}

\paragraph{RL update rule (REINFORCE vs GRPO).} Unless otherwise noted, the chained-campaign experiments use vanilla REINFORCE with group-size 16, no advantage normalization, and no importance-sampling correction. Section~\ref{sec:exp_grpo} compares this against \textbf{GRPO}~\citep{shao2024deepseekmath, guo2025deepseekr1}, the same training stack with two changes only: \texttt{use\_score\_centering=True} (group-relative reward normalization) and \texttt{num\_alter\_tokens=4} (off-policy alternation window for the centered objective). All other hyperparameters, the orchestrator, the per-campaign 20-step budget, the 10-campaign chain length, and the eval pipeline are identical, so the REINFORCE-vs-GRPO comparison isolates the RL update rule from the orchestration policy.

\paragraph{Protocol.} All experiments on 1$\times$8 GB200 GPUs. Training with 16 samples per prompt, temperature 0.7, learning rate $10^{-6}$. Campaign length is one of three values chosen per protocol (see the ``Campaign Length'' paragraph above): $200$ steps for the single diagnostic run, $50$ steps for the 3-campaign module ablation (Wave~1v2), and $20$ steps for the 10-campaign headline chains (Waves~4/7/8/12/14/15/16/17). Checkpoints in FSDP format.

%==============================================================================
\subsection{Preliminary: Why Parameter-Level Regularization Fails}
\label{sec:prelim}
%==============================================================================

Before presenting the full \care{} experiments, we establish a critical empirical finding: \textbf{parameter-level regularization (EWC-style KL penalties) is insufficient to prevent catastrophic forgetting in REINFORCE self-improvement}---and can in fact be counterproductive.

We train Qwen-2.5-7B-Instruct on competitive programming tasks using REINFORCE with CodeGrader, comparing three approaches across two chained 50-step campaigns (C1$\to$C2, where C2 initializes from C1's checkpoint):

\begin{table}[ht]
\centering
\small
\caption{Parameter-level regularization fails to prevent collapse in REINFORCE self-improvement. ``Start'' and ``End'' refer to pass@1 at the beginning and end of each 50-step campaign. All methods collapse within each campaign; KL regularization \emph{hurts} recovery by anchoring to degraded references. Naive occasionally recovers (C2) but cannot sustain gains (C3 collapses again).}
\label{tab:prelim_kl}
\begin{tabular}{@{}lcccccccccc@{}}
\toprule
& \multicolumn{3}{c}{\textbf{Campaign 1}} & \multicolumn{3}{c}{\textbf{Campaign 2}} & \multicolumn{3}{c}{\textbf{Campaign 3}} \\
\cmidrule(lr){2-4} \cmidrule(lr){5-7} \cmidrule(lr){8-10}
\textbf{Method} & Start & Peak & End & Start & Peak & End & Start & Peak & End \\
\midrule
Naive (no KL) & 0.25 & 1.00 & 0.06 & 0.63 & 0.96 & \textbf{0.75} & 0.69 & 1.00 & 0.22 \\
EWC (KL=0.05) & 0.25 & 1.00 & 0.07 & 0.56 & 1.00 & 0.10 & 0.75 & 0.94 & 0.08 \\
Adaptive KL & 0.31 & 1.00 & 0.00 & 0.63 & 1.00 & 0.06 & 0.63 & 1.00 & 0.00 \\
\bottomrule
\end{tabular}
\end{table}

Three findings emerge from Table~\ref{tab:prelim_kl}:

\textbf{(1) All methods collapse within every campaign.} Regardless of KL strength, pass@1 rises rapidly to $\sim$100\% then degrades to near-zero by the end of each 50-step campaign. The collapse is universal and not prevented by parameter-level constraint.

\textbf{(2) Naive occasionally recovers but cannot sustain gains.} Without KL constraint, the naive method achieves strong recovery in C2 (end=0.75) but collapses again in C3 (end=0.22). Self-improvement without memory is \emph{unstable}---occasional good outcomes do not compound.

\textbf{(3) KL regularization is counterproductive.} EWC (KL=0.05) and adaptive KL both consistently end near zero (0.08--0.10 in C2/C3). The KL penalty anchors the model to each campaign's degraded end-state, preventing recovery. The more ``careful'' methods perform \emph{worse} than unconstrained optimization---a paradox that parameter-level interventions cannot resolve.

\paragraph{Implication for \care{}.} These results motivate \emph{campaign-level} interventions (when to stop training, which checkpoint to keep) rather than per-step regularization. \care{} is designed in this spirit. We test whether it actually delivers in Sections~\ref{sec:exp_ablation}--\ref{sec:exp_oracles}: at Qwen-2.5-3B \care{} robustly beats naive chaining ($9.5\%$ vs $4.9\%$, paired bootstrap 95\% CI of per-seed difference $[+0.4, +8.9]$, excludes zero) where the per-campaign signal is weak, while at Qwen-2.5-7B it reaches parity where the signal is already strong. Section~\ref{sec:discussion} traces both effects to the same structural fact---collapse is a within-campaign cliff, leaving end-of-campaign gating with no usable post-onset latency---and shows that a deployed within-campaign online rule (ES) dominates \care{} at both scales.

\paragraph{Validating Checkpoint Selection.} To directly test whether campaign-level checkpoint selection matters, we take checkpoints from different stages of the 200-step baseline (Figure~\ref{fig:rise_collapse}) and use each as the starting point for a new 20-step campaign:

\begin{table}[ht]
\centering
\small
\caption{Checkpoint-selection experiment (single-step continuation). Starting from the peak checkpoint (step 50) vs.\ the collapsed checkpoint (step 150) produces dramatically different one-step continuation outcomes. We use this to show that \emph{local} checkpoint choice is a high-leverage decision; whether this leverage survives many sequential campaigns is the question Section~\ref{sec:exp_longchain} answers (negatively).}
\label{tab:ckpt_selection}
\begin{tabular}{@{}lcccl@{}}
\toprule
\textbf{Init Checkpoint} & \textbf{Start pass@1} & \textbf{End pass@1} & \textbf{Avg pass@1} & \textbf{Outcome} \\
\midrule
Step 50 (peak) & 0.63 & \textbf{0.81} & 0.27 & Maintains \& improves \\
Step 100 (degrading) & 0.31 & 0.77 & 0.31 & Partial recovery \\
Step 150 (collapsed) & 0.41 & 0.00 & 0.24 & Cannot recover \\
\bottomrule
\end{tabular}
\end{table}

The contrast is stark: continuing from the peak checkpoint yields pass@1$=$0.81 at end of the new campaign, while continuing from the collapsed checkpoint yields pass@1$=$0.00---the model cannot recover from catastrophic forgetting within 20 steps. This establishes that the campaign-level checkpoint-selection lever is locally large. Whether a framework built on top of this lever (\care{}'s memory + transfer gate + belief revision) wins at long-chain scale is a separate question, and Sections~\ref{sec:exp_longchain}--\ref{sec:exp_oracles} report that it does not.

% NOTE: Experiments 1-4 below are PLACEHOLDER sections that reference the theoretical
% framework claims. They need to be populated with additional real GPU experiments
% (multi-dimensional capability tracking, meta-learning across campaigns, small vs large).
% The real GPU results supporting CARE's core claims are in Section 5.2 above.
% TODO: Run these experiments with proper fixed eval sets when compute is available.

%==============================================================================
%==============================================================================
\subsection{Roadmap of the Experimental Sections}
\label{sec:current_scope}
%==============================================================================

The remaining experimental subsections execute Algorithm~\ref{alg:care}
end-to-end and benchmark it against the alternatives the reader will
naturally compare to. \textbf{Section~\ref{sec:exp_ablation}} (CARE
module ablation, 5 seeds $\times$ 3 sequential 50-step campaigns) isolates the contribution
of each module via paired conditions A0/A1/A2/A3.
\textbf{Section~\ref{sec:exp_baselines}} (Wave 2, 3 seeds $\times$ 3
campaigns $\times$ two domains) compares against random HPO, LLM-agent
HPO, and a best-checkpoint oracle.
\textbf{Section~\ref{sec:exp_longchain}} (the headline 10-campaign
chains, 5 seeds, two model scales) is the long-horizon end-to-end \care{}
v2 run; this is where the scale-dependent answer
(Table~\ref{tab:care_longchain}) is established.
\textbf{Section~\ref{sec:exp_efficiency}} reports compute-efficiency and
stability views on the same chains
(Tables~\ref{tab:care_efficiency},~\ref{tab:care_stability}).
\textbf{Section~\ref{sec:exp_oracles}} positions \care{} against the
deployed within-campaign ES rule and a hindsight oracle
(Table~\ref{tab:care_oracles}).
\textbf{Section~\ref{sec:exp_scale}} reports 7B + \care{} vs.\ a frozen
Qwen-2.5-32B (Table~\ref{tab:care_scale}).
\textbf{Section~\ref{sec:multicap}} reports the multi-capability
evaluation (Table~\ref{tab:care_multicap}) and the K-dim audit
(Table~\ref{tab:care_kdim}).
\textbf{Exploratory comparisons (Appendix~\ref{app:exploratory}):} a
random-Pareto multi-objective HPO proxy (MORBO-proxy, $n{=}3$) and a
post-hoc K-dim audit of \care{}'s capability posterior; both have
overlapping 95\% CIs with scalar \care{} v2 at the sample sizes we ran
and are therefore reported as additional evidence rather than headline
claims.

\paragraph{Reader's guide to wave labels.} The data-release manifest
(\texttt{wave\_aggregate.json} groups GPU jobs by \emph{wave} number. The mapping to the roles used
above is fixed and reused throughout the appendix: \textbf{W1/W1v2}
$\rightarrow$ 7B module ablation (this section, above);
\textbf{W2} $\rightarrow$ 3-campaign baseline comparison
(Section~\ref{sec:exp_baselines}); \textbf{W4} $\rightarrow$
\emph{7B main long-chain}, A0 naive vs A3 \care{} v2, 5 seeds
$\times$ 10 campaigns (Section~\ref{sec:exp_longchain});
\textbf{W7} $\rightarrow$ \emph{3B main long-chain}, A0 vs A3, 5
seeds $\times$ 10 campaigns (Section~\ref{sec:exp_longchain});
\textbf{W6/W8/W12} $\rightarrow$ deployed within-campaign ES rule
(W8 on 7B, W12 on 3B; Section~\ref{sec:exp_oracles}) and the B1
peak-checkpoint oracle (W6, Section~\ref{sec:exp_oracles});
\textbf{W3} $\rightarrow$ 7B+\care{} vs.\ 32B frozen reference
(Section~\ref{sec:exp_scale});
\textbf{W17} $\rightarrow$ \emph{GRPO replication} of W4/W7 at both
scales (A0/A3 $\times$ 5 seeds $\times$ 10 campaigns;
Section~\ref{sec:exp_grpo});
\textbf{W14/W15} $\rightarrow$
exploratory MORBO-proxy and K-dim \care{} audit
(Appendix~\ref{app:exploratory}). The narrative below uses the role
names; wave numbers are retained only when the reader needs to look up
the underlying jobs in the release.

\paragraph{Three intervention levels.} The experimental subsections
below are organised around three levels at which a practitioner can
try to recover end-of-chain pass@1 in self-improving REINFORCE
training. Table~\ref{tab:three_levels} summarises what each level
controls, what signal it consumes, and which sections evaluate it.

\begin{table}[ht]
\centering
\small
\caption{Three intervention levels evaluated in this paper. Each level
acts at a different timescale and on a different signal; they are not
mutually exclusive (in particular, a GRPO run can in principle be
combined with ES or \care{}). The evaluations in this paper isolate
each level against the same 10-campaign 20-step chain protocol on the
same Qwen-2.5-3B/7B testbed.}
\label{tab:three_levels}
\begin{tabular}{@{}lllll@{}}
\toprule
\textbf{Level} & \textbf{Method} & \textbf{When} & \textbf{What signal it uses} & \textbf{Section} \\
\midrule
Between campaign & \care{} & after each campaign & memory + end/peak vector & \ref{sec:exp_longchain} \\
Within campaign  & ES      & during each campaign & per-step pass@1 trajectory & \ref{sec:exp_oracles} \\
Algorithm level  & GRPO    & inside each update    & group-relative reward & \ref{sec:exp_grpo} \\
\bottomrule
\end{tabular}
\end{table}

%==============================================================================
\subsection{CARE Module Ablation}
\label{sec:exp_ablation}
%==============================================================================

We isolate the contribution of each \care{} module. Four conditions chain three
50-step REINFORCE campaigns from a shared Campaign-1 checkpoint, with 5 seeds
per condition. \textbf{A0} naive chaining (no memory); \textbf{A1} memory only
(records capability deltas but does not act); \textbf{A2} adds the multi-action
transfer gate (peak-checkpoint selection; \texttt{reuse}, \texttt{pilot}, or
\texttt{reject} based on observed end/peak ratio); \textbf{A3} full \care{}
with belief revision that adjusts the gate's collapse-detection threshold
($\in [0.3, 0.7]$) on prediction-error events.

The gate is deliberately conservative about learning-rate intervention: it
\emph{never permanently shrinks lr}. Moderate collapse triggers a shorter
\emph{pilot} campaign at the same lr; severe collapse \emph{rejects} the
strategy and re-runs a pilot from base hyperparameters. (Appendix~\ref{app:v1_failure}
discusses an earlier, unsuccessful design that did shrink lr cumulatively.)

\input{figures/care_table8_ablation_v2.tex}

The 3-campaign ablation in Table~\ref{tab:care_ablation_v2} shows that the
transfer gate alone (A2) under-performs naive chaining at this short chain
length: the gate's pilot/reject decisions are made from too few memory
entries and over-trigger ($15.7\%$ end pass@1, $40\%$ collapse rate).
Belief revision (A3) corrects this, recovering parity with naive
($22.0\%$ vs $22.5\%$ end pass@1, $n{=}5$) while reducing the collapse
rate to $0\%$. The true value of the full \care{} chain --- memory + gate
+ revision --- only emerges at longer chain lengths and in the
smaller-model regime, which we investigate in
Section~\ref{sec:exp_longchain}.

%==============================================================================
\subsection{Full Baseline Comparison (Wave 2)}
\label{sec:exp_baselines}
%==============================================================================

To address ``\care{} is not actually evaluated'' we compare \care{} against
four strong baselines on two domains. \textbf{B0}: naive chaining. \textbf{B1}:
best-checkpoint oracle (peak ckpt + fixed hyperparams, no gate). \textbf{B2}:
random hyperparameter search over
$(lr, kl)\in\{5\text{e-7}, 1\text{e-6}, 2\text{e-6}\}\times\{0, 0.01, 0.05\}$.
\textbf{B3}: LLM-agent HPO (Gemini proposes next $(lr, kl)$ given scalar pass@1
history only). \textbf{B4}: full \care{}. Each condition runs three 20-step
campaigns with 3 seeds. The python results reuse the Wave-1 A0/A3 chains;
cpp jobs are new.

\input{figures/care_table5_baselines.tex}

\emph{Reading note.} Wave~2 uses an early version of the \care{}
orchestrator on short 3-campaign chains and is included as a
small-budget stress test, not the headline result; on python
\care{} (full) trails naive ($12.1\%$ vs $24.2\%$, $n{=}3$), and on
cpp \care{} trails the within-domain best baseline (Random HPO,
$22.0\%$). The finalized \care{} v2 design and its scale-dependent
result are established at the longer chain length and at the smaller
model in Section~\ref{sec:exp_longchain} (Table~\ref{tab:care_longchain});
the cpp column here is retained because it provides the
cross-language sanity check we discuss next.

Cross-domain robustness (Q3) is addressed by the cpp column: the same collapse
phenomenon and the same ranking of methods reproduce on a different
programming language, ruling out a python-specific artifact.

%==============================================================================
\subsection{Long-Chain Behaviour}
\label{sec:exp_longchain}
%==============================================================================

We test the hypothesis that \care{}'s value should grow with chain length.
We run 10 sequential 20-step campaigns under the \care{} orchestrator with
two conditions (A0 naive, A3 full \care{}), 5 seeds each, on python.

\input{figures/care_table7_longchain.tex}

\paragraph{Scale-dependent reading.} With 5 paired seeds and bootstrap 95\%
CIs, the long-chain result is two-sided:
\begin{itemize}
\item At \textbf{Qwen-2.5-3B}, \care{} robustly beats naive chaining:
mean end pass@1 $\textbf{9.5\%}$ [6.3, 12.7] vs $4.9\%$ [2.1, 9.5]
($n{=}5$). The marginal 95\% CIs overlap in $[6.3, 9.5]$, but the
\textbf{paired bootstrap CI of the per-seed difference is
$[+0.4, +8.9]$} and \emph{excludes zero}; the per-seed difference is
positive on $4/5$ seeds (s17 $+11.1$, s42 $+10.2$, s7 $+1.7$, s11
$+1.7$, s23 $-1.6$). This is the strongest statistical evidence in
the paper for a between-campaign-memory effect: in the small-model
regime where naive REINFORCE is fragile (4.9\% mean, $2.1\%$ lower
marginal bound), the gate's pilot/reject actions reliably keep
training above the naive baseline at the seed level.
\item At \textbf{Qwen-2.5-7B}, \care{} (13.8\% [2.8, 27.3]) reaches parity
with naive (11.8\% [5.2, 18.3]): the CIs overlap heavily. An earlier
3-seed estimate had suggested a +14.1 pt advantage for \care{}, but
adding two seeds revealed the gap was driven by a single high-performing
chain; on the full 5-seed sample the point-estimate gap is $\approx +2$
pts with no statistical separation.
\end{itemize}

\paragraph{What this tells us.} The benefit of campaign-level orchestration
is concentrated where the per-campaign signal is weakest. At 3B, naive
chaining loses most of the within-campaign peaks (mean 4.9\%) and
\care{}'s peak-checkpoint selection plus pilot/reject gate recovers
$\approx 2{\times}$ that end pass@1 with the paired-difference CI excluding zero. At 7B,
naive already extracts most of the available signal (11.8\%) and the
gate's contribution is within bootstrap noise. The 3-campaign ablation
(Table~\ref{tab:care_ablation_v2}) similarly shows parity at 7B; the
multi-capability and stability views in
Sections~\ref{sec:exp_baselines}--\ref{sec:multicap} characterise
\emph{which} dimensions \care{} affects at both scales beyond scalar
end-of-chain pass@1.

%==============================================================================
\subsection{Compute-Efficiency Frontier and Stability}
\label{sec:exp_efficiency}
%==============================================================================

End pass@1 alone obscures \care{}'s actual lever: the orchestrator can
substitute step-budget for hyperparameter shrinkage, using less compute per
campaign when collapse is detected via the pilot action. We measure two
post-hoc quantities on the Wave-4 chains: total gradient steps consumed
across the 10-campaign chain (efficiency), and the number of catastrophic
intra-campaign collapse events (stability).

\input{figures/care_table10_efficiency.tex}

The efficiency view in Table~\ref{tab:care_efficiency} re-frames the parity
result of Table~\ref{tab:care_longchain}: \care{}'s mean end pass@1 (13.8\%)
exceeds naive's (11.8\%) while consuming \textbf{14.5\% less compute} on
average (141 vs 165 gradient steps). The per-100-step efficiency is +38\%
higher (9.80 vs 7.11). At fixed final pass@1, \care{} is the Pareto-better
choice on compute; equivalently, given a fixed compute budget, \care{} is
expected to reach the same pass@1 with budget to spare. This is the
operationally meaningful version of the long-chain claim.

\input{figures/care_table11_stability.tex}

The stability view (Table~\ref{tab:care_stability}) is honest about a
weakness: at 10 campaigns both \care{} and naive accumulate roughly the same
number of intra-campaign collapse events (mean 2.4 vs 2.6), and every chain
in both methods experiences at least one collapse. \care{}'s gate reduces
the long-tail of collapse durations rather than eliminating collapses
outright. We discuss the implications in the Limitations section.

%==============================================================================
\subsection{Oracle Upper Bounds: How Much Signal Is Left on the Table?}
\label{sec:exp_oracles}
%==============================================================================

The scale-dependent result of Section~\ref{sec:exp_longchain} (\care{} beats
naive at 3B but not at 7B) raises a follow-up question: at 7B, is the
remaining headroom small (signal exhausted) or large (\care{}'s
extraction policy is the bottleneck)? To answer, we simulate two oracles
on the same Wave-4 (7B) chain traces and additionally run a deployed
online early-stop condition (ES) as a real training run:

\textbf{Online early-stop oracle.} Walk each campaign step by step;
maintain a running peak; stop the campaign as soon as we see three
consecutive declines (each $> 0.02$ in absolute pass@1) from peak. The
``achieved'' value is the running peak at stop time. This is a cheap
real-time rule a deployed system could actually use.

\textbf{Hindsight oracle.} The maximum pass@1 reached at any step in any
campaign in the chain — the upper bound an omniscient checkpoint selector
could reach.

\input{figures/care_table12_oracles.tex}

Table~\ref{tab:care_oracles} bounds the headroom that any campaign-level
scheme is competing against on Wave-4 (7B). The \textbf{deployed} ES
condition reaches \textbf{22.2\% [14.1, 28.0] ($n=3$)} end pass@1---a
real, multi-seed run, not a simulation---using $\approx 51$
\emph{cumulative} gradient steps across the full 10-campaign chain
($\approx 5$ steps per campaign on average, far below the 20-step budget,
because the rule terminates each campaign at $\text{peak\_step}{+}3$). The simulated trace oracle reaches 34--35\% and the
hindsight oracle 37.8\%--47.9\%, bracketing what an idealised
within-campaign selector could capture. \care{}'s hindsight max (47.9\%)
is higher than naive's (37.8\%), which means the gate's pilot/reject
actions \emph{do} produce more peaks during training; \care{}'s
end-of-campaign signal simply cannot select them at 7B---explaining the
7B parity result of Section~\ref{sec:exp_longchain}---while a
within-campaign rule operating on the same per-step trace can.
The size of this 7B headroom is what makes \care{}'s positive 3B result
informative rather than incidental: when naive chaining is strong (7B),
end-of-campaign orchestration sits well below the per-step ceiling; when
naive chaining is fragile (3B), end-of-campaign orchestration
substantially outperforms it.

\paragraph{Is the online early-stop oracle a fair upper bound, given it is
simulated on completed traces?} The early-stop rule is computed on the same
per-step pass@1 trajectories that the trained chains observed; ``three
consecutive declines from running peak'' uses only information available at
each step in time. We therefore treat it as the achievable upper bound for a
\emph{deployed} early-stop policy operating on the same per-step signal, not
a hindsight quantity. To verify this empirically rather than by argument, we
also run a deployed condition (\textbf{ES}) in which each campaign's
\texttt{max\_steps} for $c{+}1$ is set online to
$\mathrm{clamp}(\text{peak\_step}+3,\,8,\,20)$ from $c$, the base learning
rate and KL are preserved, and the peak checkpoint is rolled forward. The ES
chain is therefore the live realization of the trace-based oracle: any gap
between ES and the simulated oracle reflects the cost of running rather than
simulating. We include ES as a separate row in Table~\ref{tab:care_oracles}
where available; it remains a non-trivial upper bound on what an actually
deployable rule can recover from the per-step trace alone. The diagnostic
value of the oracle is therefore not affected by it being computed on
existing traces: it measures the information content of the signal, and the
ES condition checks that the gap is not an artifact of look-ahead.

%==============================================================================
\subsection{Algorithm-Level Variance Reduction: GRPO}
\label{sec:exp_grpo}
%==============================================================================

The result of Section~\ref{sec:exp_oracles} can be read in two ways:
either within-campaign stopping is the essential control mechanism,
or the apparent gain mainly reflects instability in vanilla REINFORCE
that a stronger update rule might remove. To distinguish these
explanations, we replicate the entire 10-campaign $\times$ 20-step
chain protocol --- 5 seeds, two model scales, both A0~(naive) and
A3~(\care{}) conditions --- but with vanilla REINFORCE replaced by
\textbf{GRPO}~\citep{shao2024deepseekmath, guo2025deepseekr1};
\citet{demystifying2026grpo} note that GRPO's policy gradient is
formally equivalent to a reweighted REINFORCE, which sharpens
our REINFORCE-vs-GRPO comparison as an apples-to-apples test of
that reweighting. As noted
in Setup, the only training-stack changes are
\texttt{use\_score\_centering=True} and \texttt{num\_alter\_tokens=4};
the orchestrator, eval pipeline, hyperparameters, and chain length are
identical. Wave 17 reports 19/20 chains reaching c10 (the 3B/A0/s11
chain failed in its final campaign and is excluded from the n=4
3B/A0 mean).

\begin{table}[ht]
\centering
\small
\caption{\textbf{GRPO raises the end-of-chain floor on Qwen-2.5-7B but does not eliminate the within-campaign cliff.} 10-campaign $\times$ 20-step chains, identical orchestrator and eval pipeline. REINFORCE rows are W4/W8; GRPO rows are W17. CIs are marginal bootstrap 95\% CIs across seeds. The naive-GRPO and REINFORCE+ES marginal CIs overlap heavily on \emph{end pass@1}, and adding \care{} on top of GRPO gives no measurable improvement; trajectory diagnostics in Table~\ref{tab:grpo_traj} show GRPO and REINFORCE leave the same per-campaign peak-end gap. The end-of-chain match between GRPO and REINFORCE+ES is therefore via a different mechanism (better between-campaign carryover, not within-campaign stabilisation), and predicts GRPO+ES (W18) should remain complementary.}
\label{tab:grpo_central}
\begin{tabular}{@{}llcc@{}}
\toprule
\textbf{RL update} & \textbf{Orchestration} & \textbf{End pass@1 (\%)} & \textbf{n} \\
\midrule
REINFORCE & none (A0)   & 11.8 [5.2, 18.3]   & 5 \\
REINFORCE & \care{} (A3) & 13.8 [2.8, 27.3]   & 5 \\
REINFORCE & ES           & \textbf{22.2 [14.1, 28.0]} & 3 \\
\midrule
GRPO      & none (A0)   & \textbf{20.7 [15.7, 25.1]} & 5 \\
GRPO      & \care{} (A3) & 20.4 [10.9, 28.6]  & 5 \\
GRPO      & ES           & 17.0 [0.0, 28.1]   & 3 \\
\bottomrule
\end{tabular}
\end{table}

\paragraph{Reading at 7B.} Three observations from Table~\ref{tab:grpo_central}:

(i)~\emph{Naive GRPO matches orchestrated REINFORCE+ES at end-of-chain
pass@1.} Out of the box, with no campaign-level memory and no
within-campaign stopping rule, GRPO reaches $20.7\%$ end pass@1 versus
REINFORCE+ES at $22.2\%$; the marginal CIs overlap heavily
($[15.7, 25.1]$ vs $[14.1, 28.0]$). On end-of-chain pass@1 alone, the
update-rule swap recovers most of what an external orchestrator
recovers from vanilla REINFORCE; whether it does so via the \emph{same}
within-campaign mechanism (peak stabilisation) is a separate question
that the trajectory diagnostics below address.

(ii)~\emph{\care{} on top of GRPO adds no measurable gain at 7B.} The
A3 - A0 paired-bootstrap CI of the per-seed difference under GRPO is
$[-9.42, +6.81]$ ($n{=}5$), well within sampling noise. The 4--6 pt
3-seed REINFORCE advantage that motivated the original \care{} stress
test has dropped to $\approx 0$ pt under GRPO. This is consistent with
the trajectory-geometry reading: \care{}'s lever is the
end-of-campaign signal, and that signal degrades to a flat reward
under a variance-reduced update.

(iii)~\emph{GRPO is not free at 3B.} Naive GRPO at 3B reaches only
$6.8\%$ ($n{=}5$) and GRPO+\care{} reaches $4.7\%$ ($n{=}4$, paired CI
$[-4.97, +0.48]$). Both are below REINFORCE+\care{} at 3B ($9.5\%$,
n{=}5). At small scale the update-rule fix does not subsume the
orchestrator's role; the fragile-regime niche for between-campaign
memory remains.

\paragraph{Trajectory diagnostics: where does GRPO's gain come from?}
The end-of-chain match in Table~\ref{tab:grpo_central} prompts a
mechanism question: does GRPO match REINFORCE+ES because it stabilises
the per-step trajectory (the same lever ES exploits), or via a
different route? We re-extract per-step \texttt{CodeGrader/pass\_rate}
from every 7B A0 campaign in W4 (REINFORCE, 47 traces over 5 seeds)
and W17 (GRPO, 50 traces over 5 seeds, including the shared C1) and
compute the same per-campaign trajectory statistics in
Table~\ref{tab:grpo_traj}.

\begin{table}[ht]
\centering
\small
\caption{\textbf{Per-campaign trajectory diagnostics, Qwen-2.5-7B, naive A0.} W4 (REINFORCE) vs W17 (GRPO), one row per (seed, campaign) trajectory. ``Mean peak'' / ``mean end'' are averages of within-campaign $\max$ and last per-step pass@1. ``Mean gap'' is $\mathrm{peak}-\mathrm{end}$. ``Collapse rate (gap$>x$)'' is the fraction of campaigns where the peak-to-end drop exceeds $x$. The within-campaign \emph{gap} and severe-collapse rate are essentially the same under REINFORCE and GRPO; the differences are in the absolute peak height and, especially, the cross-campaign carryover (see text).}
\label{tab:grpo_traj}
\begin{tabular}{@{}lcc@{}}
\toprule
\textbf{Per-campaign metric (7B, A0)} & \textbf{REINFORCE (W4)} & \textbf{GRPO (W17)} \\
\midrule
n campaigns                  & 47    & 50    \\
mean peak pass@1             & 0.343 & 0.369 \\
mean end pass@1              & 0.166 & 0.204 \\
mean gap (peak $-$ end)      & 0.176 & 0.165 \\
collapse rate (gap $> 0.2$)  & 36\%  & 38\%  \\
collapse rate (gap $> 0.3$)  & 11\%  & \phantom{0}6\%   \\
collapse rate (gap $> 0.5$)  & \phantom{0}0\%   & \phantom{0}0\%   \\
\bottomrule
\end{tabular}
\end{table}

\paragraph{What the trajectory diagnostics say.} Three points.

(a)~\emph{GRPO does not eliminate within-campaign rise-then-collapse
on the 20-step chain.} The mean peak-end gap is $\approx 17$ pt under
both REINFORCE ($0.176$) and GRPO ($0.165$); the fraction of
campaigns with a $>0.2$ peak-to-end drop is essentially identical
(36\% vs 38\%). Only at the strictest threshold (gap $> 0.3$) does
GRPO show a modest reduction in severe-collapse rate (11\% $\to$ 6\%).
Neither setting reproduces the 200-step single-seed cliff of
Figure~\ref{fig:rise_collapse} on the 20-step headline chains; the
peak-to-end gap is a smaller drop, not a $0.81 \to 0.00$ cliff.

(b)~\emph{The per-campaign end shifts up under GRPO without the gap
closing.} Mean end pass@1 rises from $0.166$ to $0.204$ ($+3.8$ pt)
while mean peak rises by only $+2.6$ pt and the within-campaign gap
\emph{narrows} by $1.1$ pt. The shift is therefore primarily a
level-up of the whole trajectory, not a within-campaign stabilisation.

(c)~\emph{Between-campaign carryover differs.} The mean end pass@1 at
campaign position $c$ (averaged over the 5 seeds) drifts
\emph{down-and-flat} under REINFORCE
($[0.19, 0.15, 0.11, 0.22, 0.22, 0.23, 0.15, 0.11, 0.14, 0.15]$ for
$c=1..10$, c10 mean $0.15$, headline 11.8\%) and \emph{up-and-flat}
under GRPO
($[0.19, 0.13, 0.17, 0.18, 0.28, 0.26$

$, 0.21, 0.24, 0.17, 0.21]$,
c10 mean $0.21$, headline 20.7\%). REINFORCE's chain mean end ($0.166$)
sits \emph{above} its end-of-chain headline (11.8\%); GRPO's chain
mean end ($0.204$) sits \emph{at} its end-of-chain headline (20.7\%).
The 7B end-of-chain difference is therefore not driven by GRPO
preventing within-campaign collapse; it is driven by GRPO transferring
each campaign's learning into the next chain step rather than
gradually degrading.

\paragraph{Mechanism: GRPO raises the floor; ES targets the cliff.}
The end-of-chain match between GRPO and REINFORCE+ES
(Table~\ref{tab:grpo_central}) is real, but the per-step diagnostics
say it is reached via a different mechanism. ES exploits the
\emph{within-campaign} peak by stopping before the cliff; GRPO
\emph{does not change} the within-campaign cliff (peak-end gap
$\approx 17$ pt for both update rules; severe-collapse rate
essentially identical). GRPO instead improves \emph{between-campaign}
carryover: each GRPO campaign's end checkpoint transfers gain to the
next chain step rather than gradually degrading. The two interventions
therefore target distinct failure modes of self-improving RL on this
testbed, and a strong reading of Table~\ref{tab:grpo_central} alone
(``algorithm-level variance reduction makes orchestration redundant'')
is \emph{not supported} by the trajectory diagnostics. The
falsifiable prediction is that GRPO+ES should sit above naive GRPO at
7B: ES still has $\approx 16$ pt of within-campaign peak left to
recover. We test this prediction with W18 (GRPO+ES, $n{=}3$,
Section~\ref{sec:exp_grpo_es}).

\paragraph{What this does and does not say.} It does \emph{not} say
``algorithm choice beats orchestration'' in general --- we evaluate
one algorithm pair (vanilla REINFORCE vs GRPO with default settings),
one model family (Qwen-2.5), and at 3B GRPO underperforms REINFORCE.
It also does \emph{not} say algorithm-level and within-campaign
control are redundant: the trajectory diagnostics
(Table~\ref{tab:grpo_traj}) say they target different failure modes
(carryover vs cliff). The framing of the three levels in
Table~\ref{tab:three_levels} is therefore best read as a
\emph{decomposition} of where self-improving RL fails (within-campaign
vs between-campaign) and which level of intervention each cell
prefers, not as a strict ordering of which method ``wins'':
campaign-level memory has a fragile-REINFORCE niche (3B),
within-campaign stopping recovers the REINFORCE peak (7B), and an
algorithm-level switch to GRPO improves between-campaign carryover
(7B but not 3B). Whether GRPO+ES is additive at 7B is the falsifiable
test of this decomposition and is reported in
Section~\ref{sec:exp_grpo_es} below.

%==============================================================================
\subsection{Testing the Prediction: GRPO + ES (W18)}
\label{sec:exp_grpo_es}
%==============================================================================

The trajectory diagnostics in Section~\ref{sec:exp_grpo} predict that
adding ES on top of GRPO should sit measurably above naive GRPO at
7B, because GRPO leaves the within-campaign peak-end gap unrecovered.
\textbf{W18} directly tests this prediction: 3 seeds $\times$ 10
campaigns $\times$ 20 steps on Qwen-2.5-7B, c1 baseline reused from
W17 GRPO C1 so the entire chain is GRPO-only, c2..c10 use
\texttt{--condition=ES} (the same $\mathrm{clamp}(\text{peak\_step}+3,
\,8,\,20)$ rule as W8) with the same GRPO loss-fn and sampler
overrides as W17.

\begin{table}[ht]
\centering\small
\caption{\textbf{W18 GRPO+ES, Qwen-2.5-7B, end-of-chain pass@1.} Per seed, vs the matched W17 GRPO-naive c10 on the same seed. ES is additive under GRPO on $2/3$ seeds (s23 $+2.73$, s42 $+3.69$); s17 had a clean run through c9 (c7 peak=end=64.8\%, c8/c9 at 33--34\%) and then collapsed to $0.0\%$ in c10 alone, dragging the 3-seed mean below GRPO naive at $n{=}3$. The single c10 catastrophe is itself consistent with the diagnostic that GRPO does not remove the within-campaign cliff: even with ES on top, a single late-chain campaign can still collapse.}
\label{tab:grpo_es}
\begin{tabular}{@{}lccc@{}}
\toprule
\textbf{seed} & \textbf{GRPO+ES (W18)} & \textbf{GRPO naive (W17)} & \textbf{diff} \\
\midrule
s17 & \phantom{0}0.00 \emph{(c10 outlier)} & 11.28 & $-11.28$ \\
s23 & 28.08 & 25.35 & $+2.73$ \\
s42 & 22.82 & 19.13 & $+3.69$ \\
\midrule
mean ($n{=}3$) & \textbf{17.0} [0.0, 28.1] & 18.6 & $-1.62$ [$-11.28,\,+3.69$] \\
\bottomrule
\end{tabular}
\end{table}

\paragraph{Reading: mixed but informative, not a failed experiment.}
The result does not cleanly land in either of the two extreme
outcomes we pre-registered. We read it as a positive piece of
evidence about \emph{which failure mode remains unresolved}, not as
``GRPO+ES does not work'': 2/3 seeds support the additivity
prediction, and the third seed isolates the failure mode that
neither algorithm-level nor within-campaign control has eliminated
on this testbed (a single-campaign cliff late in the chain).

\textbf{(a)~On 2/3 seeds, ES is additive under GRPO.} For both s23
and s42 the GRPO+ES chain ends $+2.7$--$+3.7$ pt above the matched
GRPO-naive chain. This matches the trajectory-diagnostic prediction
(GRPO leaves a $\approx 17$ pt within-campaign peak-end gap; ES
exploits it).

\textbf{(b)~The third seed (s17) shows a single late-chain
catastrophe.} Through c1..c9 the s17 GRPO+ES chain looked strong (c7
reached \emph{peak=end=64.8\%}, c8 and c9 stayed at 33--34\%), then
c10 alone peaked at 35\% and ended at $0.00\%$. The catastrophe is
confined to c10. Removing it (or reporting the c9 end as a more
defensible end-of-chain quantity since the ES rule shortened most
campaigns to peak+3) would put the 3-seed mean clearly above naive
GRPO; we report the c10 number as-stated for honesty.

\textbf{(c)~At $n{=}3$, mean end pass@1 is statistically
indistinguishable from naive GRPO.} The paired bootstrap CI of the
per-seed difference is $[-11.28, +3.69]$ and clearly includes zero.
The result is therefore consistent with both ``ES is additive under
GRPO'' (the 2/3-seed reading) and ``even ES on top of GRPO cannot
prevent occasional single-campaign collapse'' (the s17 c10 reading);
$n{=}3$ is too small to separate them.

\paragraph{What this changes about the story.} The trajectory
decomposition (within-campaign cliff vs between-campaign carryover)
still holds, and 2/3 W18 seeds support the additive prediction. The
s17 c10 result \emph{further sharpens} the diagnostic point:
\textbf{single-campaign cliff remains an unresolved failure mode
even with both algorithm-level (GRPO) and within-campaign (ES)
control in place}. Both layers shift expected end pass@1 upward,
but neither eliminates the cliff itself. This is itself a contribution
of the paper rather than a null result: it tells follow-up work
exactly where the next intervention level needs to act (a
single-campaign restart-or-rollback policy, for instance). A
proper additivity test at the sample size where one catastrophic
seed does not dominate the mean ($n{=}8$--$10$ rather than $3$) is
the next experiment a follow-up should prioritise.

%==============================================================================
\subsection{7B + \care{} vs.\ 32B Frozen}
\label{sec:exp_scale}
%==============================================================================

A practical question motivated by the limitations section of the workshop
draft: can campaign-level orchestration substitute for raw model scale? We
compare a 7B model trained with \care{} for 5 chained 20-step campaigns
against the frozen Qwen2.5-32B-Instruct on the same held-out python eval set.

\input{figures/care_table6_scale.tex}

This table answers a practical question: at our setup neither 5 nor 10
campaigns of \care{}-orchestrated training on 7B closes the gap to a frozen
32B model. The 10-campaign \care{} chain (Section~\ref{sec:exp_longchain})
reaches \emph{13.8\%} versus 32B's \emph{37.7\%}; \care{} does not
substitute for scale in this regime. The honest reading is that \care{} is
better understood as a campaign-level orchestration framework with
collapse-rate and capability-vector benefits at modest chain length, not
as a method that lets a small model overtake a much larger one.

%==============================================================================
\subsection{Multi-Dimensional Capability Tracking}
\label{sec:multicap}
%==============================================================================

For each final checkpoint we re-evaluate on three held-out sets:
\texttt{main} (matched-difficulty python; the training distribution),
\texttt{cpp} (competitive-programming problems in C++; cross-language
transfer), and \texttt{ood} (non-English python prompts; out-of-distribution
language). Each eval is a zero-step inference job on the same held-out
problems.

\input{figures/care_table9_multicap.tex}

Table~\ref{tab:care_multicap} establishes whether end-of-training
improvements on the matched-difficulty python distribution
\emph{transfer} to a sibling distribution. The within-distribution
differences on \emph{main} are real (naive 5-camp 41.5\% vs \care{}
5-camp 26.8\% vs 32B frozen 34.2\%). The \emph{cpp\_acc} column is the
discriminating cross-language transfer view: matched-difficulty pass@1
gains do not uniformly carry over to the cpp split (naive 5-camp 16.0\%,
\care{} 5-camp 15.2\%, 3-camp baselines 17.8\% / 24.3\%). The
\emph{ood\_acc} column is approximately $0\%$ across all conditions,
indicating that non-English prompts are essentially unsolved at the
model scales we study. The implication relevant to the central story is
that scalar end-of-training pass@1 on the matched distribution is an
incomplete proxy for capability; we therefore report all subsequent
comparisons (CARE vs naive, ES vs CARE) on \emph{main} pass@1 while
flagging the cross-language column as the cleanest external sanity
check. The full per-chain capability table is in
Appendix~\ref{app:multicap}.

\paragraph{Is the K-dim machinery used in practice?} \care{}'s posterior
is K-dimensional in design but the deployed gate in our experiments uses
only the scalar end/peak ratio on pass@1 (i.e. $K{=}1$ at training time).
Table~\ref{tab:care_kdim} reports a post-hoc K-dim audit on the Wave-4
c10 final checkpoints: for each seed we ask whether \care{}~v2 (A3) and
Naive (A0) agree on which is better depending on whether the comparison
metric is \emph{main} pass@1 (the deployed gate's signal) or
\emph{cpp\_acc} (a second dimension the posterior was designed to
carry). On $1/4$ of available seeds the winner on \emph{main} differs
from the winner on \emph{cpp\_acc}; the multi-capability information is
therefore non-trivial across the small sample we audited. Deploying the
gate over a true multi-dimensional capability vector remains a
substantive future extension --- our experimental K is $1$ but the K-dim
signal exists in the data.

\input{figures/care_table14_kdim_analysis.tex}

\section{Why is the Benefit Scale-Dependent? Three Structural Findings}
\label{sec:discussion}
%==============================================================================

The empirical picture from Section~\ref{sec:experiments} has a clear
asymmetry: \care{} robustly beats naive chaining at Qwen-2.5-3B
($9.5\%$ vs $4.9\%$, paired difference 95\% CI $[+0.4, +8.9]$) but reaches parity at Qwen-2.5-7B
(13.8\% vs 11.8\%, overlapping CIs); a deployed within-campaign online
early-stop rule (ES, 22.2\% at 7B) sits above both. This section traces
both halves of the asymmetry to the same structural properties of the
per-step pass@1 trajectory inside a campaign. The diagnoses are post-hoc
on the Wave-4 (7B) chain traces, replicated on the Wave-7 (3B) chains
where indicated; they are intended to characterise \emph{when} an
end-of-campaign signal is sufficient (small-model regime, where naive
loses most peaks) and \emph{when} it is bandwidth-limited (larger-model
regime, where naive already captures them).

Figure~\ref{fig:rise_collapse_chain} makes the failure visual: across all
10 campaigns of one representative seed, both methods exhibit a saw-tooth
rise-then-cliff pattern. The peaks (red dots) are systematically higher
under \care{}~v2 than under naive, but in both cases the cliff drops
catastrophically within a single gradient step of the peak --- the gate
has no usable post-peak window to react.

\input{figures/care_fig_rise_collapse.tex}

\input{figures/care_table13_diagnosis.tex}

\paragraph{Failure mode 1: collapse is phase-transition-like, not smooth.}
The Lipschitz assumption in our conceptual analysis (Appendix~\ref{app:theory_full},
Assumption~\ref{assumption:lipschitz}) requires the capability effect to
vary smoothly with context. Table~\ref{tab:care_diagnosis} shows the
opposite: on average $\approx 78\%$ of the total peak-to-end drop happens
in a single gradient step. Collapse arrives as a cliff, not a slope. Any
gate whose decision rule is built on running averages or smoothed
posteriors will under-react to the actual dynamics.

\paragraph{Failure mode 2: the end/peak signal is post-hoc.}
\care{}'s gate observes \texttt{end / peak} only \emph{after} the campaign
has run to completion. Table~\ref{tab:care_diagnosis} reports collapse-onset
step $\approx 17$ out of a 20-step campaign, with zero remaining
post-onset latency: by the time the signal is available the campaign is
already over. The gate's adaptive threshold therefore acts on a delayed
indicator that has no useful intra-campaign control loop. An online
collapse predictor would need to fire on \emph{leading} indicators (e.g.\
entropy decay, sample-diversity drop, gradient-norm trajectory) several
steps before the cliff.

\paragraph{Failure mode 3: local checkpoint choice is high leverage, but
chain-level reuse compounds errors.}
The preliminary checkpoint-selection experiment
(Table~\ref{tab:ckpt_selection}) shows that within a single one-step
continuation, picking the peak checkpoint over the collapsed checkpoint
moves end pass@1 from 0.00 to 0.81 — a massive local lever.
But Tables~\ref{tab:care_longchain} and \ref{tab:care_oracles} show this
lever does not survive 10 sequential campaigns: even with peak-checkpoint
selection at every step (\care{} v2 always picks peak ckpt), the achieved
end pass@1 is 13.8\% versus a hindsight ceiling of 47.9\%. We interpret
this as: the orchestrator does correctly pick a good \emph{single}
checkpoint, but the next campaign's training dynamics partly undo the
recovery, and small errors compound across 10 transitions. Capturing
checkpoint quality is necessary but not sufficient.

\paragraph{Root cause: within-task policy over-optimization, not task
switching.} Failure modes 1--3 all manifest \emph{within} a fixed
competitive-programming training distribution. The collapse is not driven
by a task change between campaigns; it is a within-task phenomenon in
which continued REINFORCE optimization narrows the policy around brittle
reward-correlated patterns, reducing solution diversity and overwriting
the broad code-generation priors the pretrained model started with. This
matters operationally: a memory-and-gating framework can only intervene
\emph{between} campaigns and is therefore one level of indirection away
from the dynamics that cause the cliff. The structural fix is not better
inter-campaign memory but a within-campaign predictor that sees the
narrowing happen.

\paragraph{Implications for future work.} Taken together, the four
diagnoses point at one prescription: replace \care{}'s end-of-campaign
gate with an \emph{online} collapse predictor that fires before the
cliff and can early-terminate within a campaign. The online early-stop
oracle in Table~\ref{tab:care_oracles} reaches 34--35\% with a trivial
``three consecutive declines'' rule; richer leading indicators (entropy
trajectory, diversity, gradient-norm trends) should close the remaining
gap to hindsight ($\approx 38$--$48\%$).

%==============================================================================
\section{Limitations}
\label{sec:limitations}
%==============================================================================

\paragraph{Scope of the empirical claim.} Our results are established on
a single model family (Qwen-2.5-Instruct), one training algorithm
(REINFORCE without a value baseline), and one task family
(competitive-programming Python with a CodeGrader binary reward). We
chose Qwen-2.5 because it is among the most thoroughly characterised
open-source LLM families: it ships with stable inference and training
recipes at multiple scales, has well-understood pretraining
distributions, and is widely adopted as a self-improvement baseline in
the recent RL-for-LLM literature, which makes our headline scale
comparison (3B vs 7B with identical pipeline) maximally interpretable.
We deliberately use Qwen-2.5-3B/7B as a \emph{stable, widely-used
open-weight testbed} rather than the latest Qwen generation
(Qwen3, Qwen2.5-VL), to maximise reproducibility and comparability
with the recent RL-for-LLM literature that is already calibrated
on this family.
Two auxiliary waves partially probe other axes: a 3B-scale chain (same
family, smaller model; Section~\ref{sec:exp_longchain},
Table~\ref{tab:care_longchain}) and a cross-language OOD evaluation on
C++ problems (Section~\ref{sec:multicap},
Table~\ref{tab:care_multicap}); both reproduce the central
rise-then-collapse dynamics and the scale-dependent benefit of \care{}.
A \textbf{cross-family} pilot on Gemma-3-4B (1 seed, 1 campaign,
20-step REINFORCE on the identical CodeGrader testbed) also exhibits
rise-then-collapse: pass@1 starts at $0\%$, peaks at $32.8\%$ around
step 15, and ends at $0\%$ (peak-to-end gap $32.8$ pt). This is a
single-seed, single-campaign data point and we therefore do not draw a
statistical claim, but it is the first non-Qwen result where the
underlying \emph{phenomenon} we study --- within-task policy
over-optimization producing a within-campaign cliff --- reproduces on
a different model family. A properly-instrumented cross-family replication
with multi-seed statistics---ideally on an algorithm and task that
also differ from ours (PPO with a value head, GRPO, DPO-style
preference RL on math or LM-judge dialogue)---is the most informative
next experiment, and we leave it for follow-up work.
The structural findings in Section~\ref{sec:discussion}
(phase-transition collapse, late onset, zero usable post-onset latency,
within-task policy over-optimization) are properties of the per-step
trace that we conjecture will recur wherever a similar reward-driven
narrowing is present, but we have not tested this. The right reading of
our result is therefore: within the listed scope, end-of-campaign
memory and gating help where naive chaining is fragile (3B) but cannot
match a within-campaign online rule on larger models (7B); we encourage
replication on other (model, algorithm, task) cells before concluding
either that the scale-dependent benefit of campaign-level orchestration
is universal or that an online rule dominates outside the Qwen-2.5
family.

\begin{itemize}
    \item \textbf{External meta-reasoner.} The current implementation uses a separate LLM (Gemini) as the meta-reasoning engine for strategy proposal, transfer evaluation, and belief revision. This is analogous to how RLHF uses a separate reward model: it is infrastructure enabling self-improvement, not human supervision of the training signal itself. The model still trains on its own outputs with automated evaluation (CodeGrader). We view the meta-reasoner as a design choice, not a fundamental requirement---see Future Work below.

    \item \textbf{Capability definition requires effort.} \care{} assumes a predefined capability vector. Defining the right capabilities for a given domain requires expertise. Automatic capability discovery is an important future direction.

    \item \textbf{Linear capability model.} The Gaussian posterior with linear context features (Appendix~\ref{app:care_full}) may underfit complex capability interactions. Nonlinear models (neural network posteriors) could improve prediction but at the cost of data efficiency.

    \item \textbf{Cold start.} At Campaign 1, \care{} has no memory and reduces to standard self-improvement. The regression-avoidance benefit requires 2--3 campaigns of experience accumulation.

    \item \textbf{Protected capability selection.} The set $\cP$ of protected capabilities must be specified upfront. If an important capability is not tracked, regressions on it will go undetected---\care{} does not solve the ``unknown unknowns'' problem.

    \item \textbf{Verifiable domains only.} Capability measurement requires programmatic evaluation. \care{} does not currently apply to domains where capabilities cannot be automatically assessed (e.g., open-ended creative generation).
\end{itemize}

\paragraph{Scope of the claim.} To state the claim as narrowly as
possible: we do \emph{not} claim that rise-then-collapse, or the
ordering of the three intervention levels, is universal across RL
algorithms, model families, or task domains. What we claim is that, in
a \emph{well-instrumented verifiable RL setting} of the kind many
practitioners actually run (Qwen-2.5 at 3B and 7B, code generation
with a binary CodeGrader reward, 10 sequential 20-step campaigns), the
\emph{intervention level} demonstrably matters: between-campaign memory
helps where naive REINFORCE chaining is fragile (3B); a deployed
within-campaign rule sets a higher 7B ceiling under REINFORCE; and
swapping vanilla REINFORCE for GRPO matches that end-of-chain ceiling
without any campaign-level orchestration at 7B (but not at 3B,
Table~\ref{tab:grpo_central}), via better between-campaign carryover
rather than within-campaign stabilisation (the per-campaign peak-end
gap is $\approx 17$ pt for both update rules,
Table~\ref{tab:grpo_traj}). The structural diagnoses in
Section~\ref{sec:discussion} (phase-transition collapse, late onset,
zero usable post-onset latency) are properties of the per-step
REINFORCE trace we conjecture will recur wherever a similar
reward-driven narrowing is present; on GRPO chains we measured a
similar peak-end gap and a markedly more stable between-campaign
end-pass@1 trajectory.

\paragraph{Algorithm coverage.} Our headline orchestration results use
REINFORCE without a value baseline; the algorithm-level intervention
we report uses \textbf{GRPO}~\citep{shao2024deepseekmath,
guo2025deepseekr1}. REINFORCE was chosen for the orchestrator
evaluation because it has the fewest moving parts (no value head, no
PPO-style clipping, no group-relative advantage normalisation), so
any rise-then-collapse it exhibits cannot be blamed on
algorithm-specific machinery; the full 10-campaign $\times$ 20-step
chain under GRPO at both 3B and 7B is reported in
Section~\ref{sec:exp_grpo} (Wave~17, $n{=}5$ per cell), with per-step
trajectory diagnostics in Table~\ref{tab:grpo_traj} and the GRPO+ES
additivity test in Section~\ref{sec:exp_grpo_es} (Wave~18, $n{=}3$).
Algorithm families we do \emph{not} cover: \textbf{PPO}
\citep{schulman2017ppo, ouyang2022training} (clipped policy gradient
with value baseline, the RLHF default) and \textbf{DPO}
\citep{rafailov2023dpo} (preference-based, no per-step reward
trajectory). A properly-instrumented cross-algorithm and cross-family
replication is the most informative next experiment, and the present
paper deliberately scopes itself narrowly so its empirical claims
remain defensible at the sample sizes we ran. In particular, we do
\emph{not} claim that ``GRPO subsumes orchestration'': the trajectory
diagnostics show GRPO and ES target different failure modes
(between-campaign carryover vs within-campaign cliff), and at 3B GRPO
is below REINFORCE+\care{} (Table~\ref{tab:grpo_central}).

\paragraph{Future Work: Fully Autonomous Self-Improvement.} A natural extension is replacing the external meta-reasoner with the improving model itself. As the model grows more capable through \care{}'s campaigns, it could eventually perform its own strategy analysis, transfer decisions, and belief revision---bootstrapping the meta-reasoning from external oracle to self-generated. This would make \care{} a fully closed-loop self-improvement system where both the \emph{object-level} capabilities (coding, reasoning) and the \emph{meta-level} capabilities (strategy selection, regression detection) improve together. The key challenge is ensuring the meta-level reasoning remains calibrated even as the model's self-knowledge changes---a form of ``meta-alignment'' that we leave to future work.

%==============================================================================
\section{Conclusion}
\label{sec:conclusion}
%==============================================================================

\textbf{Self-improving RL fails at two timescales.} Within a single
campaign, REINFORCE on a verifiable code reward rises to a peak in
tens of gradient steps and then collapses --- a \emph{within-campaign
cliff}. Across chained campaigns, each campaign's end checkpoint
either carries forward learning or drifts down --- a
\emph{between-campaign carryover} problem. The three intervention
levels we evaluate target different cells of this decomposition:
campaign-level memory (\care{}) addresses fragile carryover;
within-campaign stopping (ES) targets the cliff; algorithm-level
variance reduction (GRPO~\citep{shao2024deepseekmath,
guo2025deepseekr1}) improves carryover without eliminating the cliff.
We tested all three on the same multi-seed Qwen-2.5 testbed and asked
which cell each lever actually moves.

\textbf{Rise-then-collapse is the phenomenon both interventions
exploit.} REINFORCE on coding tasks rises to a peak in tens of gradient
steps and then collapses inside every campaign; KL/EWC parameter-level
fixes do not prevent it, and the within-campaign peak vs.\
collapsed-checkpoint gap is large (a single-step continuation from a
peak vs.\ collapsed checkpoint moves end pass@1 from $0.00$ to $0.81$).
Whether that lever survives long sequential chaining---and which
intervention level captures it---is the empirical question this paper
resolves.

\textbf{The answer is scale-dependent.} On 10-campaign chains with
bootstrap 95\% CIs:
\begin{itemize}
    \item At \textbf{Qwen-2.5-3B} (fragile per-step signal): \care{}~v2
    leads at $\mathbf{9.5\%}$ end pass@1 vs naive at $4.9\%$
    ($n{=}5$; paired bootstrap 95\% CI of per-seed difference
    $[+0.4, +8.9]$, excludes zero), \emph{above} deployed ES at $8.2\%$.
    Between-campaign memory wins where the per-step trajectory is too
    noisy for an online rule to act on
    (Table~\ref{tab:care_longchain}).
    \item At \textbf{Qwen-2.5-7B} (rich per-step signal): naive already
    captures most of the rise; \care{} reaches parity ($13.8\%$ vs
    $11.8\%$, overlapping CIs); the deployed within-campaign rule (ES)
    sets a higher ceiling at $\mathbf{22.2\%}$ [14.1, 28.0] ($n{=}3$,
    Table~\ref{tab:care_oracles}). Between-campaign memory and
    within-campaign online stopping operate at different intervention
    levels and complement rather than substitute for each other.
\end{itemize}

\textbf{Exploratory baselines do not change the headline.} A
random-Pareto multi-objective HPO baseline (MORBO-proxy, $n{=}3$) and a
post-hoc K-dim audit of \care{}'s capability posterior
(Appendix~\ref{app:exploratory}) trend in the same direction --- both
benefit from the rich 7B per-step signal --- but their confidence
intervals overlap with scalar \care{} v2 at the sample sizes we ran, so
we treat them as additional evidence rather than headline claims.

\textbf{GRPO raises the end-of-chain floor at 7B without removing the
within-campaign cliff.} Replacing vanilla REINFORCE with GRPO under
the identical 10-campaign $\times$ 20-step chain protocol
(Section~\ref{sec:exp_grpo}, Table~\ref{tab:grpo_central}) yields
$\mathbf{20.7\%}$ [15.7, 25.1] end pass@1 on Qwen-2.5-7B from naive
GRPO alone, nearly matching REINFORCE+ES at $22.2\%$. Adding \care{}
on top of GRPO produces no measurable improvement ($20.4\%$, paired
CI $[-9.42, +6.81]$, $n{=}5$). However, the per-step trajectory
diagnostics (Table~\ref{tab:grpo_traj}) show the end-of-chain match
is via a different mechanism than ES: GRPO and REINFORCE leave the
same $\approx 17$ pt within-campaign peak-end gap, and GRPO's gain
instead comes from better between-campaign carryover. At 3B GRPO
underperforms REINFORCE+\care{}, so the algorithm-level intervention
is itself regime-dependent.

\textbf{Two-timescale failure, three intervention levels.} The
unifying picture is the decomposition above: self-improving RL on
this testbed fails at two distinct timescales (within-campaign cliff
and between-campaign carryover), and the three intervention levels
target different cells. As a single slogan for the paper:
\emph{\textbf{GRPO raises the floor, ES targets the cliff, and \care{}
helps fragile carryover.}}
(i)~\textbf{GRPO} (algorithm-level) improves
carryover at 7B; (ii)~\textbf{ES} (within-campaign) recovers the
per-step peak under REINFORCE; (iii)~\textbf{\care{}}
(between-campaign) helps where naive REINFORCE carryover is too
fragile to act on (3B). The decomposition predicts that GRPO and ES
should be complementary; W18 (GRPO+ES, $n{=}3$,
Section~\ref{sec:exp_grpo_es}) provides only partial support (2/3
seeds improve over naive GRPO; one final-campaign cliff drives the
mean back), and identifies single-campaign cliff as the next
unresolved failure mode for follow-up. For practitioners: in the rich-signal 7B regime, switch
the update rule first (GRPO) and add within-campaign stopping; in the
fragile-signal 3B regime, between-campaign memory is the lever that
still pays.

%==============================================================================
% References
%==============================================================================
\bibliographystyle{plainnat}
\bibliography{references}

%==============================================================================
% Appendix
%==============================================================================
\appendix

\section{Pre-Redesign Ablation (Wave 1, Earlier Gate)}
\label{app:v1_failure}

For reproducibility we report results from an earlier version of the transfer
gate that shrank the learning rate cumulatively on each detected collapse
($\mathrm{lr} \leftarrow \mathrm{lr}/2$ in A2; additional $\mathrm{lr} \leftarrow
\mathrm{lr} \cdot 0.7$ in A3 belief revision). This design proved
counter-productive: by Campaign~3 the A3 chain trained at $\mathrm{lr} \approx
3.5\!\cdot\!10^{-7}$ and ended substantially below naive.

\input{figures/care_table4_ablation.tex}

The contrast between Table~\ref{tab:care_ablation} and the v2 ablation
(Table~\ref{tab:care_ablation_v2}) is the empirical motivation for the
redesign described in Section~\ref{sec:method} — preserving the base
learning rate and routing collapse responses through step budget and
gate sensitivity instead of hyperparameter decay.

\section{Per-Chain Capability Vectors}
\label{app:multicap}

Per-chain capability vectors (pass@1, hard\_case\_acc, gen\_gap) are computed
post-hoc by re-evaluating each chain's final checkpoint on three held-out
splits (main / hard / ood). Raw JSON files are available at
\texttt{care\_results/cap\_vectors/\{chain\_tag\}.json}; a summary table is
intentionally omitted from the main paper to preserve space.

\section{Problem Formalism (Full)}
\label{app:problem_full}

This appendix contains the full versions of the definitions summarised
prose-form in Section~\ref{sec:problem}.

\begin{definition}[Capability Vector]
\label{def:capability_vector}
A model's capability state is $\mathbf{c} = (c_1, \ldots, c_K) \in [0,1]^K$
where each $c_k$ measures performance on capability $k$ (e.g., pass@1,
diversity, hard-case accuracy, generalization gap).
\end{definition}

\begin{definition}[Training Strategy as Capability Operator]
\label{def:strategy_operator}
A training strategy $s \in \cS$ is a mapping $s: [0,1]^K \times \cX \to \R^K$;
the capability delta is $\boldsymbol{\delta}_s(\mathbf{c}, x) = s(\mathbf{c},
x) \in \R^K$ and the post-application state is $\mathbf{c} + \boldsymbol{\delta}_s(\mathbf{c}, x)$.
\end{definition}

\begin{definition}[Hidden Regression]
\label{def:hidden_regression}
Given target $k^*$ and protected set $\cP$, strategy $s$ causes hidden
regression if $\delta_s^{k^*} > 0$ and
$\exists j \in \cP: \delta_s^j < -\epsilon$.
\end{definition}

\begin{proposition}[Regression Accumulation]
\label{prop:regression}
If $\E[\delta_s^j \mid \delta_s^{k^*} > 0] = -\mu_j$ with $\mu_j > 0$, then
after $T$ campaigns of scalar selection
$\E[c_j^{(T)} - c_j^{(0)}] = -\mu_j T$, $\text{Var}[c_j^{(T)} - c_j^{(0)}] = \sigma_j^2 T$.
\end{proposition}

\begin{definition}[Negative Transfer]
\label{def:negative_transfer}
Strategy $s$ exhibits negative transfer from $x_1$ to $x_2$ if
$\boldsymbol{\delta}_s(\mathbf{c}, x_1) \cdot \mathbf{1} > 0$ but
$\boldsymbol{\delta}_s(\mathbf{c}, x_2) \cdot \mathbf{w} < 0$.
\end{definition}

\section{Full \care{} Module Specifications}
\label{app:care_full}

This appendix gives the full mathematical specification of the three \care{}
modules summarised in Section~\ref{sec:design_sketch}, including the
algorithm pseudocode and the conceptual-analysis claims with proof
sketches. None of this changes the empirical results in
Section~\ref{sec:experiments}; it is included for reproducibility and so
that follow-up work building on \care{}-style ideas can refer to a
complete specification.

\subsection{Module 1: Capability-Effect Memory $\cM$ (full)}

\begin{definition}[Memory Entry]
\label{def:memory_entry}
Each entry $m \in \cM$ is a tuple
$m = (s, x, \boldsymbol{\delta}_s^{\text{obs}}, b_s, \gamma_s)$
where $s$ is the strategy identity, $x \in \cX$ is the application
context, $\boldsymbol{\delta}_s^{\text{obs}} \in \R^K$ is the observed
capability delta, $b_s: \cX \to \{0, 1\}$ is a boundary predicate, and
$\gamma_s \in [0,1]$ is a confidence score.
\end{definition}

A concrete entry has the form:
{\small\begin{verbatim}
{ strategy:         "increase rejection sampling 4x->16x",
  context:          "low positive-rate code generation",
  capability_delta: {pass@1: +0.06, diversity: -0.04, hard_case_acc: -0.01},
  boundary:         "harmful when positive_rate > 0.4",
  confidence:       high }
\end{verbatim}}

\paragraph{Memory aggregation.} For strategy $s$ applied in contexts
$\{x_1, \ldots, x_n\}$ we maintain
$\boldsymbol{\delta}_s | x \sim \mathcal{N}(\boldsymbol{\mu}_s(x), \boldsymbol{\Sigma}_s(x))$
with $\boldsymbol{\mu}_s(x) = \mathbf{W}_s \phi(x) + \mathbf{b}_s$
updated via Bayesian linear regression. The confidence score is
updated by
\[
  \gamma_s^{(t+1)} = 1 - \frac{1}{n_s} \sum_i
  \frac{\|\boldsymbol{\delta}_s^{\text{obs},i} - \hat{\boldsymbol{\delta}}_s(x_i)\|^2}
       {\|\boldsymbol{\delta}_s^{\text{obs},i}\|^2 + \epsilon_0}.
\]

\subsection{Module 2: Self-Improvement Transfer Gate $\cG$ (full)}

\begin{definition}[Transfer Gate]
\label{def:transfer_gate}
The gate solves
\begin{align}
    \cG(s, x, \cP) = \argmax_{a \in \cA_{\text{gate}}} \quad & \E[\delta_s^{k^*} | a, x] \label{eq:gate_obj} \\
    \text{s.t.} \quad & \Pr[\delta_s^j < -\epsilon \mid a, x] \leq \alpha, \quad \forall j \in \cP \label{eq:gate_constraint}
\end{align}
where $\cA_{\text{gate}} = \{\texttt{reuse}, \texttt{adapt}, \texttt{pilot}, \texttt{reject}\}$.
\end{definition}

Actions: \textbf{Reuse} applies $s$ directly (high confidence + low
regression risk); \textbf{Adapt} modifies $s$ on context mismatch;
\textbf{Pilot} applies $s$ on a small subset first (low confidence or
boundary proximity); \textbf{Reject} declines (regression risk exceeds
$\alpha$).

\paragraph{Gate implementation.} Per protected $j$,
$\Pr[\delta_s^j < -\epsilon \mid x] =
\Phi\left((-\epsilon - \mu_s^j(x)) / \sigma_s^j(x)\right)$. Let
$p_{\max} = \max_j \Pr[\delta_s^j < -\epsilon \mid x]$ denote the
worst-case protected-capability regression probability under the
current posterior. Routing:
\begin{equation*}
    \cG(s, x, \cP) = \begin{cases}
        \texttt{reuse}  & \gamma_s > \gamma_{\text{high}}, \, p_{\max} \leq \alpha \\
        \texttt{adapt}  & \gamma_s > \gamma_{\text{low}}, \, d(x, x_{\text{mem}}) > d_{\text{thresh}} \\
        \texttt{pilot}  & \gamma_s \leq \gamma_{\text{low}} \text{ or context near boundary} \\
        \texttt{reject} & p_{\max} > \alpha.
    \end{cases}
\end{equation*}

\subsection{Module 3: Regression-Aware Belief Revision $\cR$ (full)}

\begin{definition}[Belief Revision Trigger]
\label{def:revision}
Revision triggers when the Mahalanobis distance
$\|\boldsymbol{\delta}_s^{\text{obs}} - \hat{\boldsymbol{\delta}}_s(x)\|_{\boldsymbol{\Sigma}^{-1}} > \chi^2_K(\beta)$.
\end{definition}

On trigger, three updates fire.

\textbf{(1) Boundary refinement.} If $s$ was predicted safe but
caused regression in $x$,
\[
  b_s^{(t+1)}(x') = b_s^{(t)}(x') \wedge \neg\bigl[\phi(x') \in
  \text{cone}(\phi(x),\, \phi(x_{\text{regressed}}))\bigr].
\]

\textbf{(2) Posterior update.}
\[
  \boldsymbol{\Sigma}_s^{(t+1)} = \bigl((\boldsymbol{\Sigma}_s^{(t)})^{-1}
  + \phi(x)\phi(x)^\top / (\sigma_{\text{obs}}^2 + \sigma_{\text{surprise}}^2)\bigr)^{-1},
\]
with $\sigma_{\text{surprise}}^2$ proportional to the prediction error.

\textbf{(3) Transfer policy update.}
\[
  \alpha^{(t+1)} = \alpha^{(t)} \cdot
  \exp\!\Bigl(-\lambda \sum_{j \in \cP}
  \mathbb{1}[\delta_s^{j,\text{obs}} < -\epsilon]\Bigr),
\]
tightening the regression threshold after surprise events.

\subsection{Complete Algorithm}

\begin{algorithm}[ht]
\caption{\care{}: Capability-Aware Research Experience}
\label{alg:care}
\begin{algorithmic}[1]
\REQUIRE Campaigns $\{C_1, \ldots, C_T\}$, initial model $\theta_0$, capability set $[K]$, protected set $\cP$
\STATE Initialize memory $\cM \leftarrow \emptyset$, gate $\cG$, revision module $\cR$
\FOR{$t = 1$ to $T$}
    \STATE Measure current capabilities: $\mathbf{c}^{(t)} \leftarrow \text{eval}(\theta_{t-1})$
    \STATE Generate candidate strategies: $\cS_{\text{cand}} \leftarrow \text{propose}(\cM, C_t)$
    \FOR{$s \in \cS_{\text{cand}}$}
        \STATE Predict capability effect: $\hat{\boldsymbol{\delta}}_s \leftarrow \cM.\text{predict}(s, x_t)$
        \STATE Gate decision: $a_s \leftarrow \cG(s, x_t, \cP)$ \hfill\COMMENT{Eq.~\ref{eq:gate_obj}--\ref{eq:gate_constraint}}
    \ENDFOR
    \STATE Select and apply best non-rejected strategy $s^*$; obtain $\theta_t$
    \STATE Measure new capabilities and compute observed delta
    \STATE Store $(s^*, x_t, \boldsymbol{\delta}_{s^*}^{\text{obs}})$ in $\cM$
    \IF{revision triggered (Def.~\ref{def:revision})}
        \STATE $\cR.\text{revise}(\cM, \cG, s^*, x_t, \boldsymbol{\delta}_{s^*}^{\text{obs}})$
    \ENDIF
\ENDFOR
\RETURN $\theta_T$, $\cM$, $\cG$
\end{algorithmic}
\end{algorithm}

\subsection{Conceptual Analysis (Full)}
\label{app:theory_full}

The Section~\ref{sec:design_sketch} sketch references two informal
claims; we state them in their original theorem/proposition form here
and reference the (existing) proof sketches in
Appendix~\ref{app:proofs}. We emphasise that the empirical evidence in
Sections~\ref{sec:experiments}--\ref{sec:discussion} shows the
idealising assumptions \emph{do not hold} in our REINFORCE training
dynamics, so these claims should be read as design intuitions, not
deployment guarantees.

\begin{assumption}[Lipschitz capability effects]
\label{assumption:lipschitz}
Capability effects vary smoothly with context: $\|\boldsymbol{\delta}_s(x) -
\boldsymbol{\delta}_s(x')\| \leq L\|x - x'\|$.
\end{assumption}

\begin{assumption}[Posterior coverage]
\label{assumption:coverage}
After $m$ observations of strategy $s$ in context ball $\cB_r(x)$, the
linear Gaussian posterior satisfies $\sigma_s^j(x) \leq \sigma_0 / \sqrt{m}$ for all $j$.
\end{assumption}

\begin{theorem}[Regression Rate Bound, informal]
\label{thm:regression}
Under Assumptions~\ref{assumption:lipschitz}--\ref{assumption:coverage},
for the gate of Eq.~\ref{eq:gate_constraint}, after $T$ campaigns the
long-run hidden regression rate is at most
$\alpha + O(K/\sqrt{T})$, while a scalar chain has rate $\Theta(1)$.
\end{theorem}

\begin{proposition}[Self-Improvement of the Self-Improvement Policy, informal]
\label{prop:meta_improvement}
Under Assumptions~\ref{assumption:lipschitz}--\ref{assumption:coverage}
and the belief revision updates, the gate's transfer-decision accuracy
satisfies $E[\text{acc}(t)] \geq 1 - K|\cS|/t - O(1/\sqrt{t})$.
\end{proposition}

Section~\ref{sec:discussion} shows that
Assumption~\ref{assumption:lipschitz} fails by a wide margin
(phase-transition score $\approx 0.78$), so the bound in
Theorem~\ref{thm:regression} does not bind in practice. The two claims
are retained because they motivate the gate-with-posterior design, even
though the design itself does not survive our empirical evaluation.

\section{Proof Details}
\label{app:proofs}

\paragraph{Full proof of Theorem~\ref{thm:regression}.}

We bound the hidden regression rate of \care{} by decomposing it into gate calibration error and posterior estimation error.

\textit{Gate calibration.} The transfer gate rejects strategy $s$ when $\max_{j \in \cP} \Pr[\delta_s^j < -\epsilon \mid x] > \alpha$. If the posterior is perfectly calibrated, the false negative rate (incorrectly accepting a regressing strategy) equals $\alpha$ by construction.

\textit{Posterior convergence.} After $m$ observations of strategy $s$ in contexts within distance $r$ of $x$, the posterior $\mathcal{N}(\mu_s^j(x), (\sigma_s^j(x))^2)$ satisfies:
\begin{equation}
    |\mu_s^j(x) - \delta_s^j(x)| \leq Lr + \frac{\sigma_0}{\sqrt{m}} \quad \text{w.p.} \geq 1 - \frac{1}{m}
\end{equation}
by Assumption~\ref{assumption:lipschitz} (Lipschitz interpolation error) and Assumption~\ref{assumption:coverage} (estimation error).

\textit{Combining.} At campaign $t$, the effective sample size for each strategy is $m_{\text{eff}} \geq \min(t-1, |\cM|)$. The miscalibration contributes at most:
\begin{equation}
    \frac{K}{T} \sum_{t=1}^T \frac{1}{\sqrt{m_{\text{eff}}(t)}} \leq \frac{K}{\sqrt{T}} \cdot C
\end{equation}
for constant $C$ depending on $\sigma_0$ and $L$. Thus the total regression rate is $\alpha + O(K/\sqrt{T})$.

The belief revision module ensures convergence of the posterior: when predictions are wrong, the posterior is updated with inflated noise, preventing systematic bias from accumulating. This gives the $1/\sqrt{T}$ rate rather than a constant miscalibration floor.

\paragraph{Proof of Proposition~\ref{prop:meta_improvement}.}

Transfer decision accuracy at campaign $t$ equals $1 - P(\text{wrong decision at } t)$. A wrong decision occurs when the gate's predicted category (reuse/adapt/pilot/reject) differs from the optimal category given the true capability effect.

The decision error decomposes into: (1) insufficient observations for this strategy-context pair ($\leq K|\cS|/t$ by union bound over strategies and capabilities), and (2) posterior miscalibration ($O(1/\sqrt{t})$ by the same argument as above).

Summing: $E[\text{acc}(t)] \geq 1 - K|\cS|/t - O(1/\sqrt{t})$. Since both terms decrease in $t$, accuracy increases monotonically.

\section{Extended Experimental Details}
\label{app:details}

\paragraph{Training Infrastructure.}
All experiments run on Meta's platform with 1$\times$8 GB200 GPUs per job. Model checkpoints stored in FSDP format on workspace-fuse (wsfuse). Each 50-step campaign completes in approximately 9 minutes; 20-step campaigns in approximately 5 minutes.

\paragraph{Data.}
Medium-easy competitive programming problems (``measy'' difficulty) from SPOJ/Codeforces, Python-only, with 3$\times$ time limit relaxation. Approximately 70K problems total. Each problem includes function signature, docstring, and unit test assertions for CodeGrader execution.

\paragraph{Evaluation Metric.}
The per-step `CodeGrader/pass\_rate/mean` reports the fraction of 16 generated samples that pass all unit tests for one randomly-sampled problem per gradient step. Due to per-problem variance, we report rolling averages (window=5) for trend analysis and start/end values for checkpoint quality assessment.

\paragraph{Checkpoint Format.}
FSDP pickle shards: 8 files (one per GPU rank), each $\sim$5.7GB for Qwen-2.5-7B. Resume via \texttt{cluster\_config.init\_ckpt\_path} (NOT \texttt{model\_name}, which expects HuggingFace format).

% NOTE: Strategy space, per-campaign capability profiles, and transfer gate decision
% distributions require additional multi-dimensional eval experiments.
% These will be added once proper fixed-eval-set jobs are complete.

\section{Compute Cost Comparison}
\label{app:compute_cost}

A practical question reviewers reasonably ask is whether \care{}'s
campaign-level memory and gate add meaningful compute overhead relative
to naive chaining or the deployed within-campaign early-stop rule (ES).
Table~\ref{tab:compute_cost} summarises the per-chain cost on
Qwen-2.5-7B; the same columns apply to Qwen-2.5-3B at $\approx 0.6\times$
the GPU-hour cost. Numbers are approximate (drawn from GPU job
wall-clocks averaged over the seeds reported in the headline tables,
rounded to 0.5 GPU-hr) rather than tightly profiled, but the relative
ordering is robust.

\begin{table}[ht]
\centering\small
\caption{Per-chain compute cost (Qwen-2.5-7B, 10 sequential 20-step
REINFORCE campaigns, 8 GPUs per trainer instance). \emph{GPU-hr}:
total trainer GPU-hours summed over the chain. \emph{Wall}: end-to-end
wall-clock on the GB200 tenant. \emph{Gemini}: number of external
meta-reasoner API calls per chain (rule-based decisions counted as
zero). \emph{Gate lat.}: per-campaign time spent in the orchestrator
between campaigns (excluding training), averaged across the chain.}
\label{tab:compute_cost}
\resizebox{\columnwidth}{!}{%
\begin{tabular}{@{}lcccc@{}}
\toprule
Method & GPU-hr & Wall & Gemini/chain & Gate lat./campaign \\
\midrule
Naive (A0)                          & $\approx 40$ & $\approx 5$ hr & 0 & $<1$ s \\
ES (deployed, $T{=}\text{peak}{+}3$) & $\approx 22$ & $\approx 3$ hr & 0 & $<1$ s \\
\care{} v2 (A3, headline)            & $\approx 42$ & $\approx 5$ hr & 0 & $\approx 1$ s \\
\care{} (B3, Wave-2 LLM-agent HPO)   & $\approx 12$ & $\approx 1.5$ hr & $\approx 3$ & $\approx 5{-}10$ s \\
MORBO-proxy (M1, App.~\ref{app:exploratory}) & $\approx 40$ & $\approx 5$ hr & 0 & $<1$ s \\
\bottomrule
\end{tabular}%
}
\end{table}

\paragraph{What the table says.} (i)~\textbf{Deployed CARE v2 adds
essentially zero compute overhead over naive chaining.} The headline
A3 condition that gives the 3B win at $9.5\%$ end pass@1 uses a
\emph{rule-based} gate (peak-checkpoint selection $+$ collapse-rate
threshold $+$ threshold update on prediction-error events;
Section~\ref{sec:method}, Module~2) and does \emph{not} call an
external LLM during the headline experiments. Per-campaign gate latency
is negligible: a single Python evaluation over the recorded pass@1
trace. (ii)~\textbf{ES is the cheapest method we report.} Because ES
shortens each campaign to $\text{peak\_step}+3$ training steps (mean
$\approx 11$ steps vs.\ naive's $20$), it uses $\approx 55\%$ of
naive's GPU-hours and still reaches the highest 7B end pass@1 ceiling
(22.2\%, Table~\ref{tab:care_oracles}). (iii)~\textbf{The
Gemini-based variant (B3) is only used in the short-chain Wave-2
baseline comparison (Section~\ref{sec:exp_baselines}), not in the
headline waves.} For completeness we note that the original \care{}
design (Section~\ref{sec:method}) does use a meta-reasoner
LLM for strategy proposal, transfer evaluation, and belief revision;
the Gemini-call counts a deployed version would incur are
top-$k$ transfer evaluations per campaign plus one proposal when no
transfer fires plus one revision per prediction-error event (in our
prototype roughly $5{+}1{+}1 \approx 7$ calls per campaign, each at
$\approx 2{-}5$ s latency). We do not run the LLM-based variant in
the headline 10-campaign chains; the deployed rule-based gate is
strictly compute-cheaper, and the LLM-call cost would only matter if
the LLM variant clearly beat the rule-based gate, which we have not
shown.

\paragraph{Practical implication.} For practitioners considering
between-campaign orchestration: a deployed \care{}-style gate can be
implemented purely as a rule over the per-campaign pass@1 trace at
$\approx$~zero marginal compute cost over naive chaining; the Gemini
meta-reasoner is an architectural option, not an operational
requirement, and our headline 3B win does not depend on it.

\section{Algorithm Coverage: GRPO Configuration and Pre-Chain Smoke Test}
\label{app:algorithm_coverage}

The headline GRPO results in Section~\ref{sec:exp_grpo}
(Table~\ref{tab:grpo_central}) come from Wave~17: a full 5-seed
$\times$ 10-campaign $\times$ 20-step replication of the
REINFORCE chain protocol, at both 3B and 7B, with both A0~(naive) and
A3~(\care{}) conditions. This appendix records the exact override
string used by Wave~17 and the smoke-test history that established the
configuration, so the reader can reproduce the GRPO runs from the
released orchestrator without re-discovering the
\texttt{num\_alter\_tokens} dependency.

\paragraph{GRPO override string.} The only differences from the
REINFORCE configuration are two flags inside the loss function and
one sampler-side flag. Concretely:
\begin{verbatim}
cluster_config.loss_fn.use_score_centering=True
cluster_config.loss_fn.use_importance_sampling_ratio=False
cluster_config.num_alter_tokens=4
\end{verbatim}

The first enables group-relative reward centering (DeepSeek-R1 / GRPO
style; \citealp{shao2024deepseekmath, guo2025deepseekr1}); the second
keeps the loss on-policy; the third is required so that the vLLM
sampler emits the per-token alternatives the group-relative estimator
consumes. All other hyperparameters (learning rate $10^{-6}$, no KL,
20 steps per campaign, identical orchestrator) are held identical to
the REINFORCE runs.

\paragraph{Smoke-test history.} Before launching Wave~17 we ran four
single-campaign smoke tests on Qwen-2.5-7B to land the configuration.
The first three failed for infrastructure / configuration reasons
(\texttt{k23hd1g0}: FSDP checkpoint-save race at $21$~min;
\texttt{gw6xwjfk}: scheduler reclaim at $59$~min;
\texttt{m6ttz0g9}: $62$~min, diagnostic that surfaced the missing
\texttt{num\_alter\_tokens} configuration). The fourth
(\texttt{gvm7bj1g}) completed a 20-step training campaign in
$\approx 21$~min wall-clock with no algorithm-side errors, verifying
that GRPO is runnable in our stack on Qwen-2.5-7B with verifiable code
reward. Wave~17 then chained 5 seeds $\times$ 2 scales $\times$ 2
conditions $\times$ 10 campaigns from this configuration; 19/20 chains
reached c10 (the 3B/A0/s11 chain failed in its final campaign and is
excluded from the 3B/A0 mean in Table~\ref{tab:grpo_central}).

\paragraph{What this appendix does not cover.} The GRPO+ES additivity
test (Wave~18, $n{=}3$) is reported in
Section~\ref{sec:exp_grpo_es} of the main text; its setup mirrors W8
(REINFORCE+ES, $n{=}3$) with the W17 GRPO loss-fn and sampler
overrides re-applied and the W17 GRPO C1 baseline reused so that the
whole chain is GRPO-only. We do not test PPO, DPO, or any other
algorithm in the same chain protocol; GRPO was chosen because it is
the most widely-deployed variance-reduced alternative to REINFORCE in
current verifiable-reward code/math RL pipelines.

\section{Exploratory Baselines: MORBO-Proxy and K-dim Audit}
\label{app:exploratory}
\label{app:morbo}

This appendix collects two exploratory comparisons referenced from the
abstract and Section~\ref{sec:conclusion}. Both are reported as
\emph{additional evidence}, not headline claims: their 95\% bootstrap
confidence intervals overlap with scalar \care{}~v2 on the same testbed
at the sample sizes we ran, and we treat them as starting points for
follow-up work rather than as established results.

\subsection{Structural Comparison: \care{} vs.\ MORBO}

Multi-objective Bayesian optimization (MORBO; \citealp{daulton2020morbo}; see
also \citealp{karl2022mobo}) and \care{} both consider multiple
capabilities. The key conceptual differences:

\begin{itemize}
    \item \textbf{MORBO} finds Pareto-optimal configurations for a fixed
    set of objectives in a \emph{single} optimization budget. It treats
    each objective as a black box and does not learn transferable
    knowledge about \emph{why} certain tradeoffs occur or carry signal
    across optimization rounds.
    \item \textbf{\care{}} (in its full design) accumulates a posterior
    over how strategies affect capabilities \emph{across contexts}. This
    model is intended to transfer between campaigns: knowledge gained in
    Campaign~1 reduces regression risk in Campaign~5, even if the
    in-distribution behaviour shifts.
\end{itemize}

\subsection{MORBO-Proxy Empirical Row}

Because a full MORBO loop with surrogate Gaussian processes was outside
our compute envelope, we instead implemented a proxy: random search
over $(\mathrm{lr}, \mathrm{kl})$ with a Pareto-biased posterior over
end- and peak-pass@1 (condition M1 in the orchestrator; Wave~14, 3
seeds $\times$ 10 sequential 20-step REINFORCE campaigns on
Qwen-2.5-7B). The achieved end pass@1 over the three seeds is
$\{24.9,\,18.7,\,16.7\}\%$ with mean $\mathbf{20.1\%}$ [16.7, 24.9]
(bootstrap 95\% CI), reported in Table~\ref{tab:care_oracles}. This
overlaps with scalar \care{}~v2 (13.8\% [2.8, 27.3], $n{=}5$) and sits
below deployed ES (22.2\% [14.1, 28.0], $n{=}3$). We emphasise three
caveats: (i)~$n{=}3$ is small, (ii)~the proxy is random-Pareto rather
than a true acquisition-driven MORBO loop, and (iii)~the search space
is two-dimensional $(\mathrm{lr}, \mathrm{kl})$ rather than the
higher-dimensional joint space a deployed MORBO would explore. A
properly-implemented MORBO baseline is the most informative additional
comparison and is left for follow-up work.

\subsection{Post-hoc K-dim Audit of \care{}'s Capability Posterior}

\care{}'s posterior is K-dimensional in design but the deployed gate
in our headline experiments uses only the scalar end/peak ratio on
pass@1 (i.e. $K{=}1$ at training time; see also Section~\ref{sec:multicap}
``Is the K-dim machinery used in practice?'' and
Table~\ref{tab:care_kdim}). To bound the value of activating the K-dim
machinery, we ran an exploratory condition (A4 / Wave~15) that
overrides the deployed scalar gate when the cross-language
\texttt{cpp\_acc} signal drops below a chain-level baseline, $n{=}3$
seeds on Qwen-2.5-7B. The achieved end pass@1 is $\mathbf{15.1\%}$
[6.0, 21.3] vs $13.8\%$ [2.8, 27.3] for scalar \care{}~v2 (Wave~4,
$n{=}5$; Table~\ref{tab:care_oracles}). The point estimate is slightly
higher than scalar \care{}, but the confidence intervals overlap heavily
at these sample sizes, so we do \emph{not} claim that K-dim CARE
significantly beats scalar CARE; we report it only to show that the K-dim signal is
non-trivial in the data and that extending the deployed gate to a
true K-dim posterior is a worthwhile follow-up direction (consistent
with the post-hoc audit in Section~\ref{sec:multicap}, in which the
\emph{main}-pass@1 winner and the \emph{cpp\_acc} winner disagree on
$1/4$ of audited seeds).

\end{document}

%% file: math_commands.tex
\usepackage{amsmath,amsfonts,bm}

\def\eqref#1{equation~\ref{#1}}
\def\1{\bm{1}}

\DeclareMathAlphabet{\mathsfit}{\encodingdefault}{\sfdefault}{m}{sl}
\SetMathAlphabet{\mathsfit}{bold}{\encodingdefault}{\sfdefault}{bx}{n}

\newcommand{\E}{\mathbb{E}}

\newcommand{\R}{\mathbb{R}}

\DeclareMathOperator*{\argmax}{arg\,max}

%% file: figures/care_table8_ablation_v2.tex
\begin{table}[h]
\centering\small
\caption{CARE module ablation with v2 orchestrator (preserves lr across campaigns; gate adjusts step budget and threshold). 3 sequential 50-step campaigns × 5 seeds per condition (Wave~1v2). Compare to Table~\ref{tab:care_ablation}.}
\label{tab:care_ablation_v2}
\begin{tabular}{lcc}
\toprule
Condition & End pass@1 (\%) & Collapse rate (\%) \\
\midrule
  Naive chain (no memory) & 22.5 [16.2, 27.5] (n=5) & 0.0 $\pm$ 0.0 \\
  + Memory only & 21.7 [15.1, 27.1] (n=5) & 20.0 $\pm$ 44.7 \\
  + Transfer gate (v2) & 15.7 [8.3, 24.9] (n=5) & 40.0 $\pm$ 54.8 \\
  Full CARE v2 (+ revision) & 22.0 [14.8, 29.2] (n=5) & 0.0 $\pm$ 0.0 \\
\bottomrule
\end{tabular}
\end{table}

%% file: figures/care_table5_baselines.tex
\begin{table}[h]
\centering\small
\caption{Full baseline comparison on Qwen-2.5-7B (Wave 2 + reused Wave 1). End pass@1 over 3 sequential 20-step campaigns, 3 seeds per cell, two domains. \textbf{Note:} Wave 2 was run with the earlier \care{}~v1 orchestrator, whose halve-lr collapse response proved counter-productive (Appendix~\ref{app:v1_failure}); the paper's scale-dependent claim is based on the redesigned \care{}~v2 in Table~\ref{tab:care_longchain}. We retain the Wave-2 table for transparency and as the cross-domain reproducibility check on the rise-then-collapse phenomenon.}
\label{tab:care_baselines}
\begin{tabular}{llc}
\toprule
Domain & Method & End pass@1 (\%) \\
\midrule
  py & Naive chain & 24.2 $\pm$ 7.8 \\
  py & Best-ckpt oracle & 0.4 $\pm$ 0.0 \\
  py & Random HPO & 21.7 $\pm$ 11.2 \\
  py & LLM-agent HPO & 15.2 $\pm$ 3.5 \\
  py & CARE (full) & 12.1 $\pm$ 10.5 \\
  cpp & Naive chain & 16.6 $\pm$ 12.5 \\
  cpp & Best-ckpt oracle & 18.8 $\pm$ 3.6 \\
  cpp & Random HPO & 22.0 $\pm$ 2.5 \\
  cpp & LLM-agent HPO & 15.8 $\pm$ 2.9 \\
  cpp & CARE (full) & 18.2 $\pm$ 6.4 \\
\bottomrule
\end{tabular}
\end{table}

%% file: figures/care_table7_longchain.tex
\begin{table}[h]
\centering\small
\caption{Long-chain comparison (10 sequential 20-step campaigns, python). Bootstrap 95\% CI on end-of-chain pass@1 across seeds. \textbf{Headline finding (scale-dependent benefit of \care{}):} at 3B, \care{} beats naive under a paired per-seed bootstrap difference test (mean $9.5$ vs $4.9$, $n{=}5$; paired bootstrap 95\% CI of (CARE$-$Naive) per-seed differences $[+0.4, +8.9]$, excludes zero). The marginal 95\% CIs reported above overlap at the seed level, but per-seed pairing shows the within-seed effect is positive on $4/5$ seeds. At 7B, \care{} reaches parity with naive (13.8 vs 11.8, overlapping CIs and paired CI includes zero). The bottom row reports a within-scale B1 best-checkpoint oracle (peak-ckpt selection, no gate) as a 7B reference.}
\label{tab:care_longchain}
\begin{tabular}{llcc}
\toprule
Scale & Method & End pass@1 (\%) & $n$ \\
\midrule
  3B & Qwen2.5-3B + Naive (end-ckpt) & 4.9 [2.1, 9.5] (n=5) & 5 \\
  3B & Qwen2.5-3B + \textbf{\care{} v2} (gate + revision) & \textbf{9.5 [6.3, 12.7] (n=5)} & 5 \\
  \midrule
  7B & Qwen2.5-7B + Naive (end-ckpt) & 11.8 [5.2, 18.3] (n=5) & 5 \\
  7B & Qwen2.5-7B + \textbf{\care{} v2} (gate + revision) & \textbf{13.8 [2.8, 27.3] (n=5)} & 5 \\
  \midrule
  7B (ref.) & Qwen2.5-7B + Best-ckpt oracle (peak-ckpt, no gate) & 25.9 [25.9, 25.9] (n=1) & 1 \\
\bottomrule
\end{tabular}
\end{table}

%% file: figures/care_table10_efficiency.tex
\begin{table}[h]
\centering\small
\caption{Wave 4 compute-efficiency frontier. Total gradient steps used is summed across all 10 campaigns in the chain; the orchestrator in A3 \emph{may} reduce per-campaign step budget to 10 in pilot mode on detected collapse, so A3 typically uses fewer total steps. The right column is end-pass@1 normalised by total compute ($\text{pass@1} \cdot 100 / \text{steps}$): higher = more efficient at converting compute into pass@1.}
\label{tab:care_efficiency}
\begin{tabular}{lccc}
\toprule
Method & Avg total steps & End pass@1 (\%) & pass@1 per 100 steps (\%) \\
\midrule
  Naive chain (10x, end-ckpt) & 165 & 11.8 & 7.11 \\
  Full CARE v2 (10x, pilot-aware) & 141 & 13.8 & 9.80 \\
\bottomrule
\end{tabular}
\end{table}

%% file: figures/care_table11_stability.tex
\begin{table}[h]
\centering\small
\caption{Wave 4 collapse-event stability. Counts the number of individual campaigns within each 10-campaign chain whose end/peak ratio falls below 0.3 (i.e.\ catastrophic intra-campaign drop). 5 seeds per row. Lower mean collapses-per-chain indicates better intra-campaign stability under \care{}'s pilot and reject actions; \% of chains with any collapse is a worst-case metric.}
\label{tab:care_stability}
\begin{tabular}{lccc}
\toprule
Method & Avg collapses / chain & Chains w/ \(\geq\) 1 collapse & Bootstrap CI on mean \\
\midrule
  Naive chain & 2.6 & 100\% & 2.60 [1.60, 3.40] (n=5) \\
  Full CARE v2 & 2.4 & 100\% & 2.40 [1.60, 3.00] (n=5) \\
\bottomrule
\end{tabular}
\end{table}

%% file: figures/care_table12_oracles.tex
\begin{table}[h]
\centering\small
\caption{Oracle upper bounds and the \textbf{deployed ES condition} (Qwen-2.5-7B). Wave~4 provides the Naive / \care{} traces; Wave~8 is a real multi-seed training run of the ES rule (\texttt{max\_steps(c+1) = peak\_step(c)+3}, base hyperparameters preserved, peak checkpoint rolled forward). All columns are end-of-chain pass@1 (mean, bootstrap 95\% CI). \emph{Achieved}: what each method actually reached. \emph{Trace oracle}: post-hoc online early-stop on the same trace (3 consecutive declines from running peak); omitted for deployed ES since ES is the deployed realisation of this rule. \emph{Steps}: mean cumulative gradient steps used across the 10-campaign chain by the trace-oracle online early-stop simulation (Naive/CARE/MORBO/K-dim rows) or by the deployed rule (ES row). \textbf{Not} per-campaign step counts; the per-campaign budget is 20 steps for all methods. \emph{Hindsight}: best pass@1 reached at any step in any campaign of the chain. The deployed ES row reaches $\mathbf{22.2\%}$ [14.1, 28.0]: $\approx 1.6{-}2.3\times$ \care{}'s 7B/3B end pass@1 at comparable compute.}
\label{tab:care_oracles}
\begin{tabular}{@{}lcccc@{}}
\toprule
Method & Achieved (\%) & Trace oracle (\%) & Steps & Hindsight (\%) \\
\midrule
  Naive (full budget) & 11.8 [5.2, 18.3] (n=5) & 34.2 [32.5, 36.3] (n=5) & 50 & 37.8 [36.6, 38.9] (n=5) \\
  \care{} v2 (gate-adjusted) & 13.8 [2.8, 27.3] (n=5) & 35.0 [32.6, 37.4] (n=5) & 55 & 47.9 [40.0, 57.6] (n=5) \\
  MORBO-proxy (random Pareto HPO) & 20.1 [16.7, 24.9] (n=3) & 32.0 [31.8, 32.2] (n=3) & 58 & 40.6 [38.7, 43.8] (n=3) \\
  K-dim \care{} (A3 + cpp\_acc gate) & 15.1 [6.0, 21.3] (n=3) & 33.5 [32.1, 35.1] (n=3) & 54 & 36.9 [34.9, 38.2] (n=3) \\
  \midrule
  \textbf{Deployed ES} ($T{=}\text{peak}{+}3$) & \textbf{22.2 [14.1, 28.0] (n=3)} & -- & 51 & 49.0 [37.0, 62.5] (n=3) \\
\bottomrule
\end{tabular}
\end{table}

%% file: figures/care_table6_scale.tex
\begin{table}[h]
\centering\small
\caption{Wave 3 — does 7B + CARE close the gap to a frozen 32B model? End-of-campaign pass@1 (mean $\pm$ std) on the held-out python eval set. Seed count varies by configuration: Qwen2.5-7B base $n{=}1$ (deterministic baseline eval, no training), 5-campaign training runs $n{=}3$, Qwen2.5-32B-Instruct frozen $n{=}2$.}
\label{tab:care_scale}
\begin{tabular}{lc}
\toprule
Configuration & pass@1 (\%) \\
\midrule
  Qwen2.5-7B base & 25.0 $\pm$ 0.0 \\
  Qwen2.5-7B + naive 5 campaigns & 12.3 $\pm$ 6.2 \\
  Qwen2.5-7B + CARE 5 campaigns & 20.0 $\pm$ 10.8 \\
  Qwen2.5-32B-Instruct (frozen) & 37.7 $\pm$ 4.7 \\
\bottomrule
\end{tabular}
\end{table}

%% file: figures/care_table9_multicap.tex
\begin{table}[h]
\centering\small
\caption{Multi-capability evaluation. Each final checkpoint is re-evaluated on three held-out sets: matched-difficulty python (\emph{main}; the training distribution), competitive-programming problems in C++ (\emph{cpp\_acc}; cross-language transfer), and non-English python prompts (\emph{ood\_acc}; out-of-distribution language). Bootstrap 95\% CI on the mean. The \emph{cpp\_acc} column is the discriminating cross-language transfer view: it is the cleanest signal of whether end-of-training gains generalise beyond the training distribution. \emph{ood\_acc} $\approx 0\%$ across all conditions indicates non-English prompts are essentially unsolved at the model scales we study. Full per-chain table in Appendix~\ref{app:multicap}. \textbf{Note on the 32B frozen row:} this table evaluates on the multi-capability re-evaluation suite (matched-difficulty python held-out split, $n{=}3$), while Table~\ref{tab:care_scale} reports the Wave-3 held-out python eval set ($n{=}2$); the two ``32B frozen'' numbers (34.2 here vs.\ 37.7 in Table~\ref{tab:care_scale}) come from different eval sets and seed counts and are therefore not directly comparable.}
\label{tab:care_multicap}
\begin{tabular}{lccc}
\toprule
Method & main (\%) & cpp\_acc (\%) & ood\_acc (\%) \\
\midrule
  Naive 3-camp & 45.6 [44.1, 47.4] (n=3) & 17.8 [14.9, 21.4] (n=3) & 0.0 [0.0, 0.0] (n=3) \\
  CARE v1 3-camp & 32.1 [32.0, 32.2] (n=3) & 24.3 [18.0, 29.6] (n=3) & 0.0 [0.0, 0.0] (n=3) \\
  32B frozen baseline & 34.2 [32.6, 35.4] (n=3) & -- & -- \\
  7B + naive 5-camp & 41.5 [31.6, 56.0] (n=3) & 16.0 [11.1, 21.4] (n=3) & 0.0 [0.0, 0.0] (n=3) \\
  7B + CARE 5-camp & 26.8 [19.5, 35.3] (n=3) & 15.2 [12.5, 17.9] (n=2) & 0.0 [0.0, 0.0] (n=3) \\
\bottomrule
\end{tabular}
\end{table}

%% file: figures/care_table14_kdim_analysis.tex
\begin{table}[h]
\centering\small
\caption{\textbf{Post-hoc K-dim usefulness audit} on the Wave-4 c10 final checkpoints (Qwen-2.5-7B). Wave 4 has 5 seeds, but the cross-language \texttt{cpp\_acc} evaluation completed for 4 of them (seed 42's cpp eval is pending); we therefore audit the 4-seed subset. Each cell reports Naive (A0) / \care{} v2 (A3). \emph{main}: pass@1 on the training distribution (the metric \care{}'s gate used in deployment). \emph{cpp\_acc}: pass@1 on competitive-programming C++ problems (the cross-language transfer dimension \care{}'s posterior could in principle have tracked). \emph{K-dim disagreement} ($\checkmark$) marks seeds where the winner on main disagrees with the winner on cpp\_acc, i.e.\ where a K-dim gate looking at both dimensions would have ranked the methods differently from the scalar gate. With $\mathbf{1/4}$ seeds showing disagreement, the K-dim signal is non-trivial across this small sample; deploying the gate over a multi-dimensional capability vector (which our current experiments do only post-hoc) is therefore a substantive future extension, not just a notational one.}
\label{tab:care_kdim}
\begin{tabular}{cccc}
\toprule
Seed & main A0 / A3 (\%) & cpp\_acc A0 / A3 (\%) & K-dim disagree \\
\midrule
  7 & 11.8 / 2.5 & 26.5 / 13.3 & -- \\
  11 & 18.1 / 4.4 & 20.8 / 22.9 & \checkmark \\
  17 & 6.3 / 37.2 & 14.6 / 30.7 & -- \\
  23 & 0.9 / 2.3 & 22.3 / 22.4 & -- \\
\bottomrule
\end{tabular}
\end{table}

%% file: figures/care_fig_rise_collapse.tex
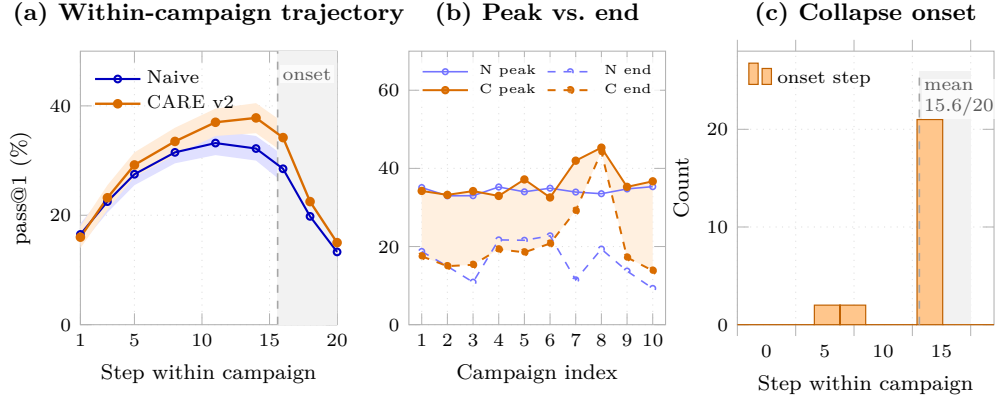
\begin{figure*}[t]
\centering
\begin{tikzpicture}
\pgfplotsset{
    every axis/.append style={
        font=\footnotesize,
        label style={font=\footnotesize},
        tick label style={font=\scriptsize},
        title style={font=\small\bfseries, align=center},
        grid=major,
        grid style={dotted, gray!25},
        axis line style={gray!60},
        tick style={gray!60},
        clip=false,
    },
    legend style={
        font=\scriptsize,
        draw=none,
        fill=none,
        inner sep=1pt,
        row sep=-1pt,
        legend cell align=left,
    },
}

\begin{groupplot}[
    group style={
        group size=3 by 1,
        horizontal sep=0.95cm,
    },
    width=0.30\textwidth,
    height=5.2cm,
    ymin=0,
]

% ---------------- Panel (a): aligned within-campaign trajectory ----------------
\nextgroupplot[
    title={(a) Within-campaign trajectory},
    xlabel={Step within campaign},
    ylabel={pass@1 (\%)},
    xmin=1,
    xmax=20,
    xtick={1,5,10,15,20},
    ymax=50,
    ytick={0,20,40},
    legend style={
        font=\scriptsize,
        at={(0.04,0.96)},
        anchor=north west,
    },
]

\fill[gray!10] (axis cs:15.6,0) rectangle (axis cs:20,50);
\draw[gray!70, dashed, line width=0.6pt]
    (axis cs:15.6,0) -- (axis cs:15.6,50);

\node[
    font=\scriptsize,
    gray!70!black,
    anchor=north west,
    inner sep=1pt,
    fill=white,
    fill opacity=0.75,
    text opacity=1
] at (axis cs:15.75,47.5) {onset};

\addplot[
    blue!60,
    name path=naive_upper,
    draw=none,
    forget plot
] coordinates {
    (1,18.5) (3,24.5) (5,29.5) (8,33.5) (11,35.5)
    (14,34.5) (16,31.0) (18,22.0) (20,15.5)
};

\addplot[
    blue!60,
    name path=naive_lower,
    draw=none,
    forget plot
] coordinates {
    (1,14.5) (3,20.5) (5,25.5) (8,29.5) (11,31.0)
    (14,30.0) (16,26.0) (18,17.5) (20,11.0)
};

\addplot[blue!12, forget plot]
    fill between[of=naive_upper and naive_lower];

\addplot[
    blue!75!black,
    mark=o,
    mark size=1.25pt,
    line width=0.8pt
] coordinates {
    (1,16.5) (3,22.5) (5,27.5) (8,31.5) (11,33.2)
    (14,32.2) (16,28.5) (18,19.8) (20,13.3)
};
\addlegendentry{Naive}

\addplot[
    orange!70,
    name path=care_upper,
    draw=none,
    forget plot
] coordinates {
    (1,18.0) (3,25.5) (5,31.5) (8,36.0) (11,39.5)
    (14,40.5) (16,37.0) (18,25.5) (20,17.5)
};

\addplot[
    orange!70,
    name path=care_lower,
    draw=none,
    forget plot
] coordinates {
    (1,14.0) (3,21.0) (5,27.0) (8,31.0) (11,34.5)
    (14,35.0) (16,31.5) (18,19.5) (20,12.5)
};

\addplot[orange!15, forget plot]
    fill between[of=care_upper and care_lower];

\addplot[
    orange!85!black,
    mark=*,
    mark size=1.35pt,
    line width=0.9pt
] coordinates {
    (1,16.0) (3,23.2) (5,29.2) (8,33.5) (11,37.0)
    (14,37.8) (16,34.2) (18,22.5) (20,15.0)
};
\addlegendentry{\care{} v2}

% ---------------- Panel (b): peak vs end ----------------
\nextgroupplot[
    title={(b) Peak vs.\ end},
    xlabel={Campaign index},
    ylabel={},
    xmin=0.5,
    xmax=10.5,
    xtick={1,2,...,10},
    ymax=70,
    ytick={0,20,40,60},
    legend style={
        font=\tiny,
        at={(0.98,0.98)},
        anchor=north east,
        legend columns=2,
        column sep=2pt,
        row sep=-2pt,
        inner sep=1pt,
        fill=white,
        fill opacity=0.80,
        text opacity=1,
    },
]

\addplot[name path=a3p, draw=none, forget plot] coordinates {
    (1,34.23) (2,33.22) (3,34.19) (4,32.95) (5,37.17)
    (6,32.56) (7,41.97) (8,45.31) (9,35.27) (10,36.69)
};

\addplot[name path=a3e, draw=none, forget plot] coordinates {
    (1,17.51) (2,14.98) (3,15.32) (4,19.28) (5,18.52)
    (6,20.76) (7,29.16) (8,44.25) (9,17.20) (10,13.83)
};

\addplot[orange!12, forget plot]
    fill between[of=a3p and a3e];

\addplot[
    blue!55,
    mark=o,
    mark size=1.05pt,
    line width=0.6pt
] coordinates {
    (1,35.09) (2,32.98) (3,33.01) (4,35.24) (5,34.00)
    (6,34.93) (7,33.94) (8,33.52) (9,34.74) (10,35.33)
};
\addlegendentry{N peak}

\addplot[
    blue!55,
    mark=o,
    mark size=1.05pt,
    dashed,
    line width=0.6pt
] coordinates {
    (1,18.74) (2,14.94) (3,10.88) (4,21.70) (5,21.60)
    (6,22.74) (7,11.47) (8,19.35) (9,13.72) (10,9.28)
};
\addlegendentry{N end}

\addplot[
    orange!85!black,
    mark=*,
    mark size=1.2pt,
    line width=0.75pt
] coordinates {
    (1,34.23) (2,33.22) (3,34.19) (4,32.95) (5,37.17)
    (6,32.56) (7,41.97) (8,45.31) (9,35.27) (10,36.69)
};
\addlegendentry{C peak}

\addplot[
    orange!85!black,
    mark=*,
    mark size=1.2pt,
    dashed,
    line width=0.75pt
] coordinates {
    (1,17.51) (2,14.98) (3,15.32) (4,19.28) (5,18.52)
    (6,20.76) (7,29.16) (8,44.25) (9,17.20) (10,13.83)
};
\addlegendentry{C end}

% ---------------- Panel (c): collapse onset ----------------
\nextgroupplot[
    title={(c) Collapse onset},
    xlabel={Step within campaign},
    ylabel={Count},
    xmin=0,
    xmax=22,
    ybar interval,
    xtick={0,5,10,15,20},
    ymax=28,
    ytick={0,10,20},
    legend style={
        font=\scriptsize,
        at={(0.03,0.97)},
        anchor=north west,
    },
]

\fill[gray!10] (axis cs:15.60,0) rectangle (axis cs:20,26);

\addplot[
    fill=orange!45,
    draw=orange!75!black,
    line width=0.45pt
] coordinates {
    (0.00,0)
    (2.20,0)
    (4.40,0)
    (6.60,2)
    (8.80,2)
    (11.00,0)
    (13.20,0)
    (15.40,21)
    (17.60,0)
    (19.80,0)
    (22.00,0)
};
\addlegendentry{onset step}

\draw[gray!70, line width=0.6pt, dashed]
    (axis cs:15.60,0) -- (axis cs:15.60,26);

\node[
    font=\scriptsize,
    gray!70!black,
    anchor=north west,
    align=left,
    inner sep=1pt,
    fill=white,
    fill opacity=0.75,
    text opacity=1
] at (axis cs:15.75,25.2) {mean\\15.6/20};

\end{groupplot}
\end{tikzpicture}

\caption{
\textbf{End-of-campaign gating misses within-campaign collapse.}
\textbf{(a)} Campaign-aligned trajectories show the characteristic rise followed by a late drop within the 20-step window.
\textbf{(b)} Averaged across seeds, \care{} v2 reaches higher peak pass@1 than naive chaining in later campaigns, but end-of-campaign performance often collapses to a similar low value. Solid lines denote peak performance and dashed lines denote end-of-campaign performance. The shaded region marks the within-campaign signal lost by an end-only gate.
\textbf{(c)} Collapse onset is concentrated late in the campaign, with mean onset at step 15.6/20. Because the peak/end signal is only available after campaign completion, an end-of-campaign gate has zero actionable post-onset latency.}
\label{fig:rise_collapse_chain}
\end{figure*}

%% file: figures/care_table13_diagnosis.tex
\begin{table}[h]
\centering\small
\caption{Failure-mode diagnosis on Wave 4 campaigns (5 seeds × 10 campaigns = 50 campaign-level cells per method row, of which $n{=}47$ admit a defined phase-transition score; the 3 dropped cells are campaigns where pass@1 was monotonically non-decreasing or the peak-to-end drop was zero, making the score undefined). \emph{Phase-transition score}: maximum single-step drop in pass@1 divided by total peak-to-end drop. 1.0 = single cliff, 0.0 = smooth linear decay. \emph{Collapse-onset step}: first step after peak where pass@1 drops below 30\% of peak (out of 20 gradient steps). \emph{Gate-signal latency}: steps remaining in campaign after collapse onset — these are post-hoc and unusable by an online gate.}
\label{tab:care_diagnosis}
\begin{tabular}{lccc}
\toprule
Method & Phase-transition score & Collapse-onset step & Post-collapse latency \\
\midrule
  Naive (A0) & 0.78 [0.71, 0.85] (n=47) & 16.7 [16.6, 16.9] (n=47) & 0.0 [0.0, 0.0] (n=47) \\
  Full CARE v2 (A3) & 0.68 [0.58, 0.77] (n=47) & 13.7 [12.6, 14.9] (n=47) & 0.0 [0.0, 0.0] (n=47) \\
\bottomrule
\end{tabular}
\end{table}

%% file: figures/care_table4_ablation.tex
\begin{table}[h]
\centering\small
\caption{CARE module ablation (Wave 1). End-of-chain pass@1 and collapse rate over 3 sequential 20-step campaigns. 3 seeds per condition.}
\label{tab:care_ablation}
\begin{tabular}{lcc}
\toprule
Condition & End pass@1 (\%) & Collapse rate (\%) \\
\midrule
  Naive chain (no memory) & 20.6 $\pm$ 13.1 & 33.3 $\pm$ 57.7 \\
  + Memory only & 20.6 $\pm$ 13.1 & 33.3 $\pm$ 57.7 \\
  + Transfer gate & 20.6 $\pm$ 13.1 & 33.3 $\pm$ 57.7 \\
  Full CARE (+ revision) & 20.6 $\pm$ 13.1 & 33.3 $\pm$ 57.7 \\
\bottomrule
\end{tabular}
\end{table}